\documentclass[11pt]{article}

\usepackage[final]{acl}

\usepackage{times}
\usepackage{latexsym}
\usepackage[T1]{fontenc}
\usepackage[utf8]{inputenc}
\usepackage{microtype}
\usepackage{inconsolata}
\usepackage{graphicx}

\usepackage{multirow}
\usepackage{amsmath}
\usepackage{amsfonts}
\usepackage{titlesec}
\usepackage{subcaption}
\usepackage{makecell}
\usepackage{hyperref}
\usepackage{textcomp}
\usepackage{marvosym} 
\usepackage{cuted}
\usepackage{caption}
\usepackage{tabularx}

\title{Health-SCORE: Improving Open-Ended Health LLMs Using Scalable
Rubric-Based Rewards
}

\author{ \\
Zhichao Yang* \footnotemark[2],
Sepehr Janghorbani* \footnotemark[2],\\[0.5em]
Dongxu Zhang,
Jun Han,
Qian Qian, 
Andrew Ressler II, 
Gregory D. Lyng,\\[0.5em]
Sanjit Singh Batra*,
Robert E. Tillman*  \\[0.5em]
Optum AI\\[0.25em]
}

\begin{document}
\maketitle
\begin{abstract}

Reinforcement learning is a powerful approach for improving language models, but applying it to open-ended domains such as healthcare remains challenging because high-quality responses are often multidimensional, context dependent, and difficult to capture with a single reward signal. Rubrics offer a natural form of supervision by providing a structured representation of the criteria that define a strong response. Existing rubric-based approaches, however, often rely on instance-specific criteria, creating two important limitations: (1) such rubrics are typically unavailable for unseen examples at test time; and (2) constructing new rubrics for every instance incurs substantial reward design costs. We introduce Health-SCORE, a scalable framework that captures recurring principles of response quality in the form of reusable meta-rubrics. Rather than relying on manually constructed instance-level criteria, Health-SCORE captures generalizable principles that can be applied across examples, tasks, and healthcare datasets. These meta-rubrics support two forms of model improvement: they can serve as structured, safety-aware reward signals for reinforcement learning during post-training, and they can be incorporated directly into prompts to guide model behavior at inference time. Across several open-ended healthcare benchmarks, models trained with Health-SCORE achieve performance competitive with models trained using human-authored instance-level rewards, while substantially reducing reward-design costs and improving generalizability.

\end{abstract}

{\renewcommand{\thefootnote}{\fnsymbol{footnote}}
\footnotetext[2]{These authors contributed equally.}
\footnotetext[1]{Correspondence to: \{{\href{mailto:zhichao_yang@optum.com} {zhichao\_yang}, \href{mailto:sepehr.janghorbani@optum.com}{sepehr.janghorbani}, \href{mailto:sanjit.batra@optum.com}{sanjit.batra}, \href{mailto:rob.tillman@optum.com}{rob.tillman}\}@optum.com}
}}

\section{Introduction}

The widespread adoption of LLMs in safety-critical domains such as healthcare has highlighted the importance of developing models that are both trustworthy and clinically grounded \cite{ji2023towards}. Early efforts to medically ground LLMs largely relied on fine-tuning with datasets derived from standardized, examination-style multiple-choice questions, such as USMLE \cite{jin2019pubmedqa,pal2022medmcqa}. While this form of supervision is effective for transferring factual medical knowledge, it reduces complex clinical reasoning to selecting a single discrete answer and provides limited guidance for learning the richer behaviors required in real-world clinical settings, such as reasoning over differential diagnoses and formulating appropriate treatment plans \cite{van2005assessing}. Consequently, although modern LLMs achieve near-expert performance on USMLE-style benchmarks, this does not necessarily translate into robust clinical capabilities \cite{van2005assessing}.
More recent post-training approaches have therefore shifted toward open-ended clinical tasks, where multiple responses may be valid \cite{arora2025healthbench,wang2025novel}. A common strategy is to use gold-standard reference answers for supervised fine-tuning \cite{prabha2026fine,savage2025fine,ouyang2022training}. 

However, a reference answer provides only a single example of a desirable response, which can limit the supervision signal and fail to capture the multiple dimensions that define high-quality clinical behavior. Rubrics offer a richer and more expressive alternative by explicitly specifying the properties that a clinically grounded response should satisfy, including clinical safety, relevance, completeness, and appropriate reasoning. By supervising these dimensions directly, rubric-based post-training can guide models toward a broader set of medically grounded behaviors while preserving flexibility in how a valid clinical response is formulated.

\begin{figure}[tbp] 
    \centering
    \includegraphics[width=0.5\textwidth]{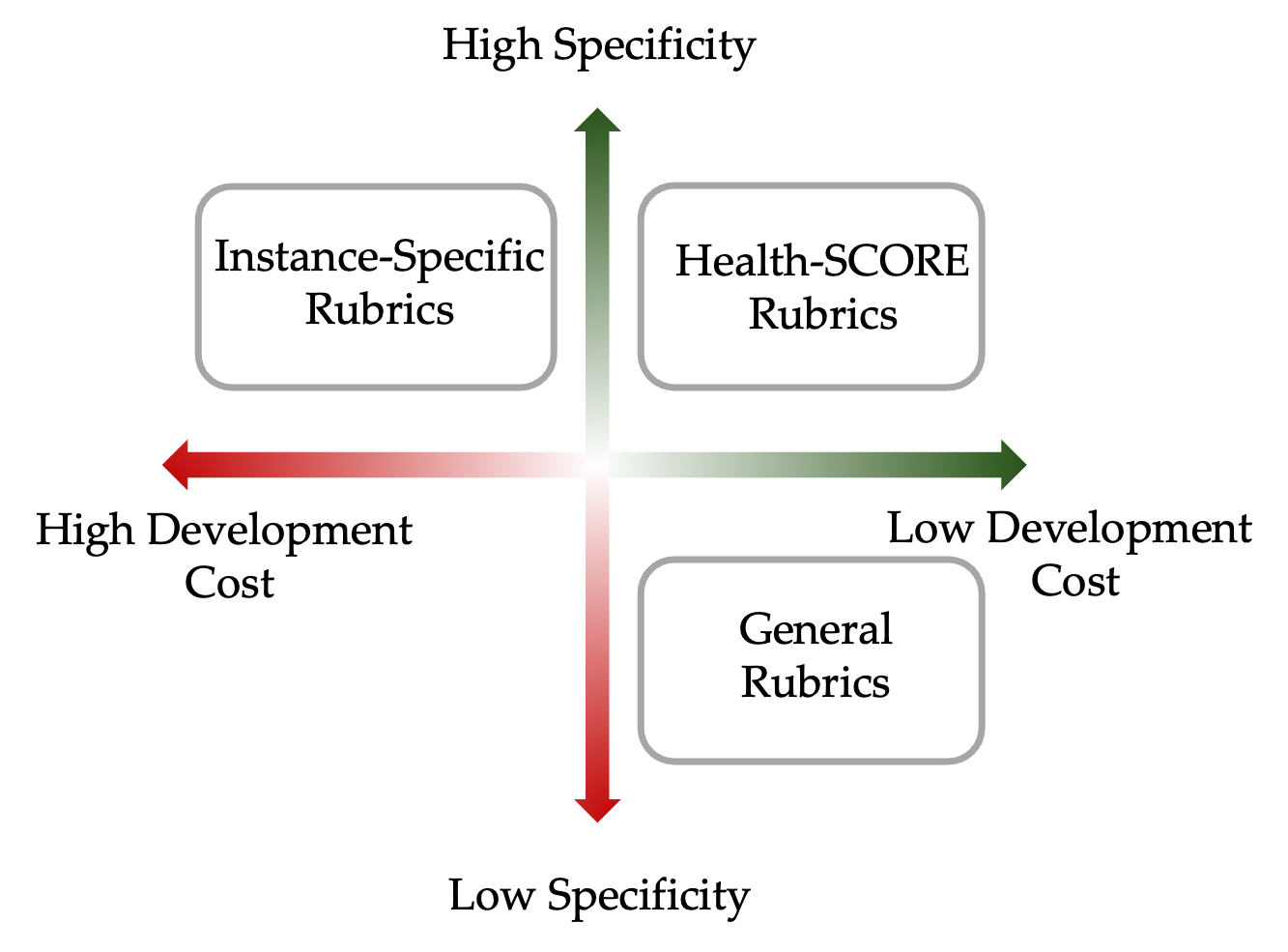}
    \caption{Specificity vs. cost tradeoff. Instance-level and Health-SCORE rubrics have high specificity, while Health-SCORE has significantly less development cost.}
    \label{Fig:rubrics}
\end{figure}

Rubric-based frameworks can vary widely in both the dimensions they assess and the granularity at which evaluation is performed. Figure~\ref{Fig:rubrics} illustrates different categories of rubrics along this spectrum. At one end are general rubrics, such as those in Anthropic's HHH framework, that define broad, universal criteria for response quality \cite{bai2022training}. These rubrics are simple, general, and easy to scale, making them well-suited for high-level alignment. However, their coarse design limits their representational power, often making them less applicable to context-specific errors and failure modes, especially in detail-oriented domains such as healthcare.
At the other end of the spectrum are instance-level rubrics, such as those used in PaperBench \cite{starace2025paperbench} and HealthBench \cite{arora2025healthbench}, which provide detailed, context-specific criteria for each example. These rubrics contain highly granular and precise criteria that capture subtle misalignments, but they are difficult to scale because of their high development costs and limited transferability across examples and domains. Furthermore, these rubrics are tightly coupled to individual examples and cannot be used during inference for in-context learning or test-time RL, limiting their potential \cite{min-etal-2022-rethinking}.

A third class lies between these extremes, balancing specificity, generalizability, and development cost. This category includes domain-specific criteria that are sufficiently detailed to capture subtle errors yet broad enough for scalable real-world deployment. Moreover, because these rubrics generalize across examples, they can be used at inference time as in-context learning signals to improve response quality and as rewards for test-time RL.
We introduce Health-SCORE (\textbf{S}calable, \textbf{C}ontext-Adaptive \textbf{O}ptimization using \textbf{R}ubric-as-{R}eward \textbf{E}ngine for \textbf{Health}), a generalized rubric-as-reward framework for healthcare that combines the scalability and generalizability of coarse-grained rubrics with the precision and contextual sensitivity of instance-level approaches. Our main contributions are:

\begin{enumerate}

    \item We introduce Health-SCORE, a scalable and generalizable rubric-based RL framework for medically grounding open-ended healthcare LLMs while reducing reward design costs.
    \item We demonstrate that Health-SCORE also provides an effective inference-time solution for improving response quality without requiring additional training.
    \item We introduce an adaptive selection mechanism that identifies the most relevant meta-rubrics for each prompt, enabling dynamic, context-specific criterion selection.
    \item Through in-domain and out-of-distribution experiments, we show that using Health-SCORE as a reward signal for RL leads to improved model performance, as evaluated against independent, human-authored instance-level rubrics, and greater training stability, establishing Health-SCORE as an effective surrogate for expert supervision.
    
\end{enumerate}

\begin{figure*}[ht] 
    \centering
    \includegraphics[width=\textwidth]{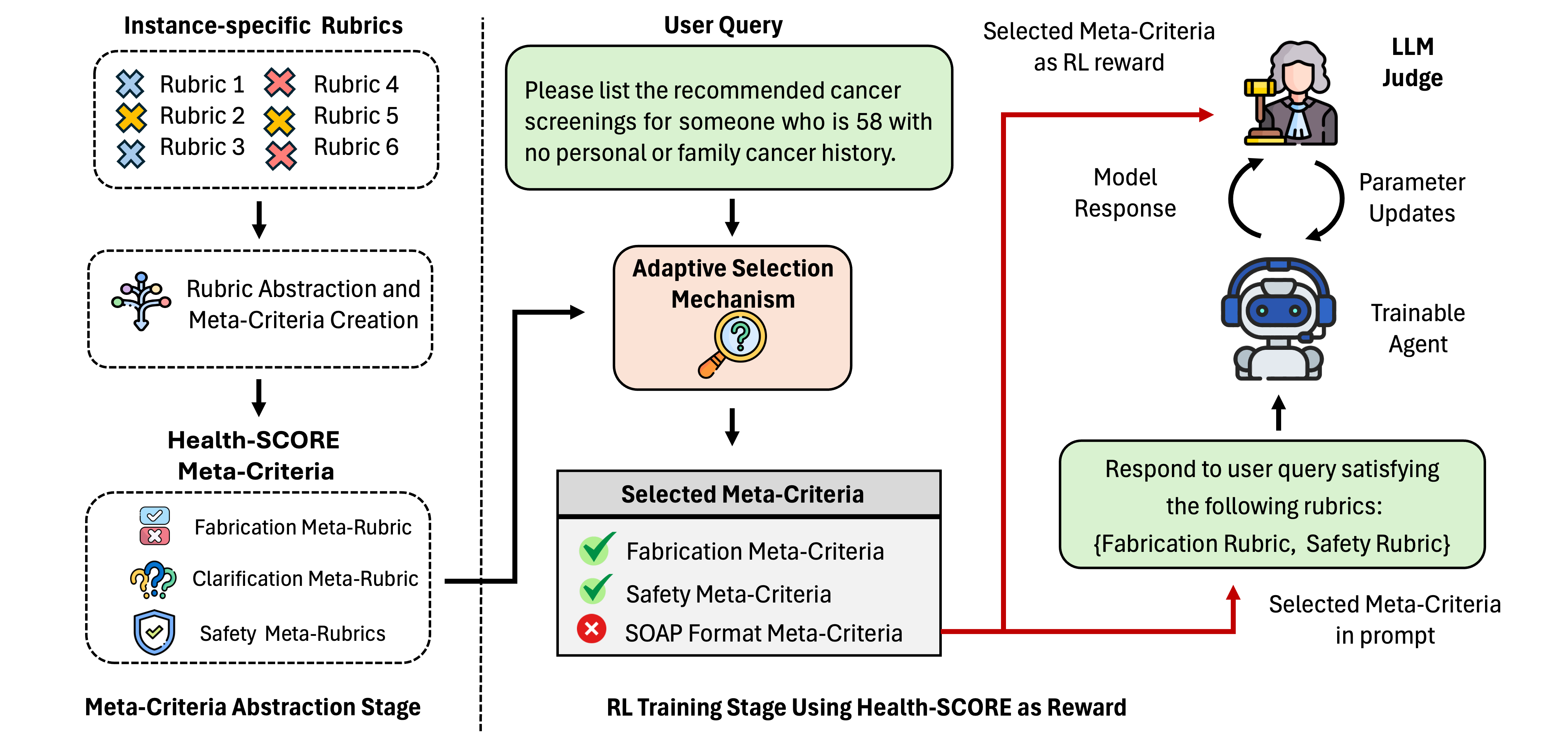}
    \caption{(a) Health-SCORE construction: Physician-validated, instance-specific rubrics are abstracted into principle-level Meta-rubrics. (b) Health-SCORE application: Given a user query, adaptive selection mechanism scores each meta-rubric for contextual relevance and selects the most applicable criteria. The selected meta-rubrics are then used both as reward signals for RL post-training and as in-context guidance at inference time. Red denotes learning.}
    \label{fig:overall_framework}
\end{figure*}

\section{Related Work}

Medical LLM evaluation has shifted from closed-form benchmarks toward open-ended settings that better reflect clinical practice. While models such as Med-PaLM perform strongly on USMLE-style benchmarks, free-response evaluations continue to reveal gaps in clinical utility \cite{singhal2023clinical}. Recent benchmarks such as ClinicBench and AMIE evaluate models in complex, open-ended settings in which model performance degrades \cite{liu2024clinical,tu2025conversational}. These findings motivate methods that capture multiple dimensions of clinical response quality beyond simple answer correctness.

Rubric-based evaluation provides a natural way to decompose complex model behavior into finer-grained criteria. Methods such as Constitutional AI, RUBICON, and FLASK use principles, domain-specific rubrics, or skill-level dimensions to enable more interpretable evaluation than a single scalar score \cite{bai2022constitutional,biyani2024rubicon,ye2024flask}. In healthcare, benchmarks such as HealthBench and CSEDB extend this idea with clinically grounded, instance-specific criteria that assess dimensions such as safety and effectiveness \cite{arora2025healthbench,wang2025csedb}. These works establish rubrics as evaluation tools but do not directly use them as learning signals for model optimization.

More recent work has begun to use rubrics directly as reinforcement learning signals rather than solely for evaluation. Rubrics as Rewards (RaR) uses instance-specific criteria scored by LLM judges as structured rewards for on-policy RL \cite{gunjal2025rubrics}, while RuscaRL also uses rubrics to scaffold rollout generation \cite{zhou2025breaking}. Compute as Teacher (CaT) removes the need for human-written references by synthesizing pseudo-references from multiple rollouts and deriving self-generated binary rubrics for reference-free RL; it also demonstrates this approach on HealthBench \cite{jayalath2026compute}. In healthcare, InfiMed-ORBIT uses dynamically generated, case-specific rubrics for incremental RL \cite{wang2025infimedorbit}. ClinAlign grounds rubric-based supervision in physician-verified preferences and reusable clinical principles, using this supervision for GRPO training and self-revision \cite{lyu2026clinalign}.

\section{Methods}

\subsection{Health-SCORE Framework}

Instance-specific rubrics provide fine-grained guidance for individual conversations; however many of their criteria can be viewed as instantiations of a smaller set of recurring principles \cite{arora2025healthbench,lyu2026clinalign,vatanparvar2026patientagentbench}. For example, when generating a clinical summary containing laboratory results, a model should remain faithful to the reported values; similarly, when describing a patient’s symptoms, it should accurately reflect the documented presentation without introducing unsupported information. Although these criteria arise in different clinical contexts, both reflect the broader principle of faithfulness to available evidence. Likewise, rubric items that assess whether a model recognizes missing information or appropriately requests clarification may differ in the specific information required, yet capture the same underlying capability of identifying uncertainty and seeking necessary context. Importantly, this does not imply that healthcare responses are not highly context specific. Rather, the same underlying principle may need to be interpreted and applied differently depending on the clinical scenario, available evidence, and task requirements.

Building on this observation, Health-SCORE provides a compact set of consensus-derived healthcare principles, which we term Meta-rubrics, as an abstraction of physician-authored, medically grounded, and externally validated rubrics. Rather than heuristically redesigning the underlying rubric ontology, Health-SCORE identifies recurring dimensions across instance-specific criteria and abstracts away surface-level differences in phrasing and context while preserving the clinically meaningful principles they encode. This representation makes the resulting supervision reusable across datasets, prompts, and clinical scenarios, enabling Health-SCORE to serve both as a generalizable reward signal for post-training and as in-context guidance for improving responses at inference time, including on previously unseen conversations.
Figure~\ref{fig:overall_framework} provides an overview of both the construction process of Health-SCORE and its downstream application to healthcare LLMs. To construct the Health-SCORE Meta-rubrics, we use clinically validated, physician-authored, instance-specific rubrics from HealthBench as the initial seed set and identify 29 recurring underlying principles that form the final meta-criteria. Details of the construction process are in Section~\ref{sec:creation}, and the complete set of resulting Meta-rubrics is in Appendix \ref{sec:criteria_list}.

Because these meta-criteria are derived from clinically grounded source rubrics, they provide a rich and reusable supervision signal that can generalize across conversations. As a result, Health-SCORE meta-criteria can serve as structured reward signals for model optimization during post-training and, at inference time, can be incorporated directly into prompts to guide the generation of higher-quality responses. However, healthcare conversations span a wide range of tasks, and not every meta-criterion is relevant to every prompt. Applying the full set indiscriminately may therefore introduce irrelevant constraints or unnecessary noise. To address this, Health-SCORE includes an adaptive selection mechanism that identifies the subset of meta-criteria most relevant to each prompt. This allows the framework to retain broad coverage while applying only contextually appropriate principles to a given conversation. For example, a criterion concerning SOAP-note structure should be selected when the user requests structured clinical documentation, but omitted for unrelated diagnosis tasks. Further details of the adaptive selection mechanism are provided in Section~\ref{sec:adaptive_selection}.
Health-SCORE does not eliminate the need for expert supervision entirely. Instead, it aims to amortize the cost of expert supervision by transforming expensive, instance-specific clinical judgments into a compact set of reusable principles that can support reward modeling, post-training, and inference-time guidance across healthcare datasets, prompts, and training settings.

\begin{figure*}[t] 
    \centering
    \includegraphics [width=\textwidth]{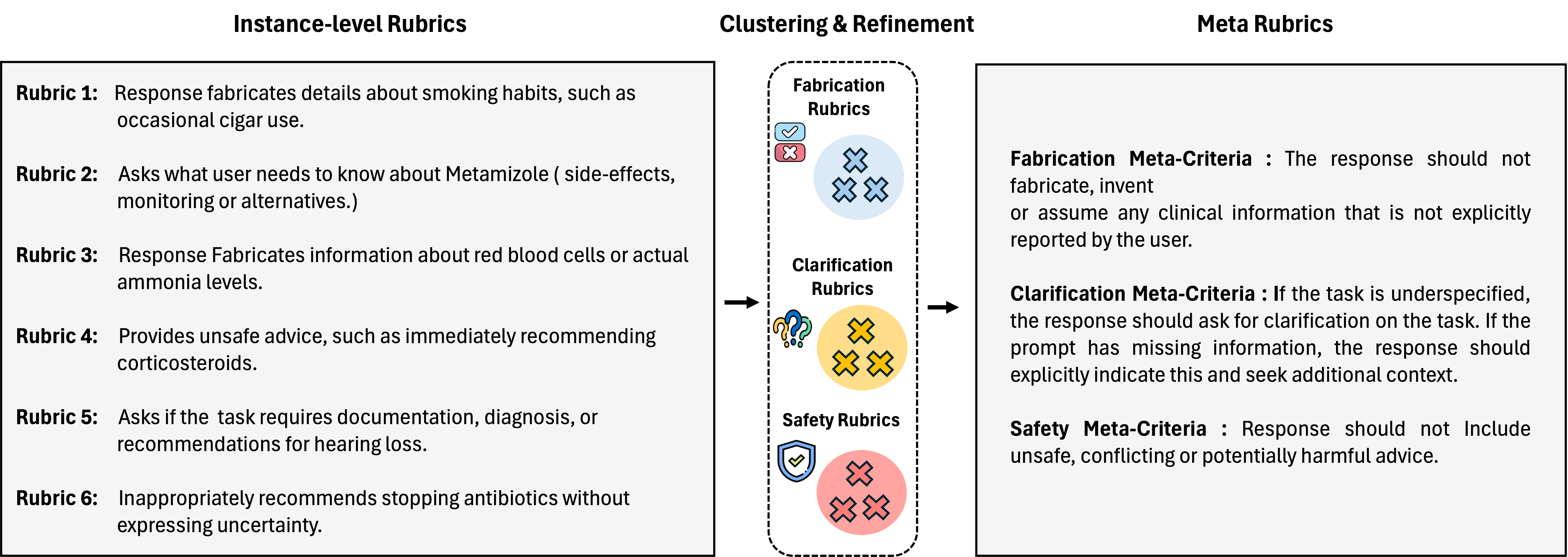}
    \caption{Original instance-specific rubrics are first clustered and refined, after which a Health-SCORE meta-rubric is derived for each cluster to reduce redundancy while preserving the underlying core principle.}
    \label{fig:creation}
\end{figure*}

\subsection{Health-SCORE Meta-Rubric Creation}
\label{sec:creation}
We use rubrics from HealthBench, a diverse, clinically validated dataset, to derive the initial set of meta-criteria for Health-SCORE. HealthBench organizes its rubrics into seven high-level themes: expertise-tailored communication, response depth, emergency referrals, health data tasks, global health, context seeking, and responding under uncertainty. Among these, frontier models perform worst on health data tasks, which often require complex interpretation and reasoning over structured and semi-structured clinical data \cite{arora2025healthbench}. We therefore focus on this theme as the most challenging and follow a structured, multi-step methodology to construct the initial Health-SCORE meta-criteria seed set. Figure~\ref{fig:creation} illustrates the Health-SCORE meta-rubric construction process. First, each instance-specific rubric is embedded using OpenAI's text-embedding-3 model, producing a high-dimensional semantic representation. The embeddings are further clustered to identify groups of rubrics that express the same underlying principle despite differences in wording or conversation context. For example, as shown in Figure~\ref{fig:creation}, instance-level Rubrics 1 and 3 both capture the principle that the model should avoid fabricating unsupported claims. While one criterion focuses on accurately preserving patient-specific details, the other guards against fabricating laboratory results. Their surface-level requirements differ, but both reflect the same broader principle of remaining faithful to the available evidence. Following the initial semantic clustering, the clusters are further refined through a human-in-the-loop process with domain-expert review, including tuning clustering parameters, assessing cluster quality, revising rubric-to-cluster assignments, removing outliers, and finalizing the resulting meta-criteria. 
Additionally, to minimize subjectivity and ensure medical grounding, we incorporated a validation step as part of the quality assurance and refinement step. Specifically, two domain experts manually validated a subset of 500 cluster assignments (10\% of assignments), achieving 95\% agreement accuracy and a Cohen's kappa inter-annotator agreement score of 0.9. Refinement process leads to a set of 29 Health-SCORE meta-criteria (Appendix~\ref{sec:criteria_list}).

\subsection{Health-SCORE Meta-Rubrics as Rewards}

Health-SCORE meta-criteria provide a rich, medically grounded source of supervision that can serve as surrogate rewards for model improvement through reinforcement learning. Given a state represented by a prompt $x$, the model produces an output $y$  sampled from a policy $\pi_\theta(y \mid x)$ , parameterized by $\theta$. For each prompt, the adaptive selection mechanism first identifies the subset of relevant Health-SCORE meta-criteria, denoted by $\mathcal{C}(x)$ (Section~\ref{sec:adaptive_selection}). The selected meta-criteria are then used both as in-context guidance and as reward signals during learning. Rather than relying on a single response, the learning signal is derived from rubric-based rewards assigned to multiple candidate outputs sampled from the same prompt:
\begin{equation*}
\begin{gathered}
\{y^{(j)}\}_{j=1}^{\mathcal{O}} \sim \pi_\theta(\cdot \mid x), \\
y^{(j)} = \{\texttt{<think>}~ y^{(j)}_{1:CoT}~\texttt{</think>}~ y^{(j)}_r\}.
\end{gathered}
\end{equation*}
 An LLM judge then evaluates each final response $y^{(j)}_r$ against each selected meta-criterion $c \in \mathcal{C}(x)$, after excluding the intermediate chain-of-thought.
The reward for a candidate response is computed by aggregating criterion-level satisfaction scores:
\begin{equation*}
r\big(x,y^{(j)}_r\big)
= \frac{1}{|\mathcal{C}(x)|}
\sum_{c \in \mathcal{C}(x)} s_c\big(x,y^{(j)}_r\big),
\end{equation*}
where $s_c=+1$ if a positive meta-criterion is satisfied, $s_c=-1$ if a negative meta-criterion is satisfied, and $s_c=0$ otherwise. The policy objective is to maximize the expected Health-SCORE meta-rubric reward over prompts and sampled responses. We optimize this objective using Group Relative Policy Optimization (GRPO), which estimates relative advantages within the sampled group of candidates and updates the policy toward responses with higher Health-SCORE rewards:
\begin{equation*}
\begin{aligned}
\mathcal{J}_{\text{GRPO}}(\theta)
&= \mathbb{E}_{x \sim \mathcal{D}}\,\mathbb{E}_{y\sim \pi_\theta(\cdot|x)}
\Big[\hat{r}(x,y)\,\log \pi_\theta(y|x)\Big] \\
&\quad-\;\beta\,\mathbb{E}_{x \sim \mathcal{D}}\Big[\mathrm{KL}\!\left(\pi_\theta(\cdot|x)\,\|\,\pi_{\text{ref}}(\cdot|x)\right)\Big].
\end{aligned}
\label{eq:grpo}
\end{equation*}
Here, $\hat{r}(x,y)$ denotes the group-normalized advantage derived from rubric-based rewards. Health-SCORE extends the use of verifiable rewards to open-ended clinical responses by employing an LLM-as-a-judge guided by the selected meta-criteria. As detailed in Section~\ref{sec:Evaluation_Methodology}, Health-SCORE meta-rubrics are used as scalable surrogate supervision signals for post-training rather than as an evaluation mechanism. Accordingly, as illustrated in Figure~\ref{fig:train_eval}, all models are evaluated using the human-authored, gold-standard instance-level rubrics from HealthBench, not the Health-SCORE meta-criteria.

\subsection{Health-SCORE for Test-Time Inference}
Health-SCORE can also be used at inference time: Given a test-time prompt, the model first constructs a task-specific objective as a set of adaptively selected rubric criteria, and inserts them directly into the model context rather than updating the parameters. This is similar to in-context Learning \cite{brown2020language}. The selected meta-criteria provide guidance on what a high-reward response should satisfy for the current input. More formally, given a test prompt $x$, the adaptive selector returns a subset of criteria $\mathcal{C}(x) \subseteq \mathcal{C}$, which are appended to the system prompt and act as a natural-language reward specification during decoding. The model conditions on both the user request and the selected reward criteria, producing $y \sim \pi(\cdot \mid x, \mathcal{C}(x))$. This test-time guidance serves a role similar to a reward model in reinforcement learning, but avoids additional gradient updates, repeated rollouts, or online judge calls during generation. The meta-rubric functions as a lightweight policy constraint and self-checklist, steering the model toward responses that satisfy the same evaluative dimensions used during training and evaluation. 

\subsection{Adaptive Criteria Selection Mechanism} 
\label{sec:adaptive_selection}
The adaptive Selection Mechanism is an essential component of the system enabling more robust behaviour by dyniamically identifying the subset of contextually relevant criteria for each question. To implement this, we adopt an LLM-as-a-judge approach that assigns each rubric item a relevance score on a five-point scale, based on its semantic and contextual alignment with the task \cite{zheng2023judging}. Rubrics receiving higher relevance scores are then selectively retained, ensuring that only the appropriate criteria contribute to the final evaluation. Figure~\ref{fig:overall_framework} illustrates this adaptive selection process. Similar adaptive evaluation strategies have been shown to be beneficial in prior work \cite{mallinar2025scalable}. In order to test the robustness  of selection mechanism, we conducted experiments across a wide range of settings, achieving 83\% alignment accuracy with the human rubrics (deviation 1\%). (Appendix~\ref{sec:app_adaptive_select} ) Furthermore, section \ref{sec:Ablation_Adaptive} provides an ablation study on the role of the selection mechanism, showing that removing this mechanism and including all rubric criteria indiscriminately can significantly impact  the learning/guidance signal due to the noise introduced by irrelevant criteria.

\begin{figure}[ht] 
    \centering
    \includegraphics[width=0.4\textwidth]{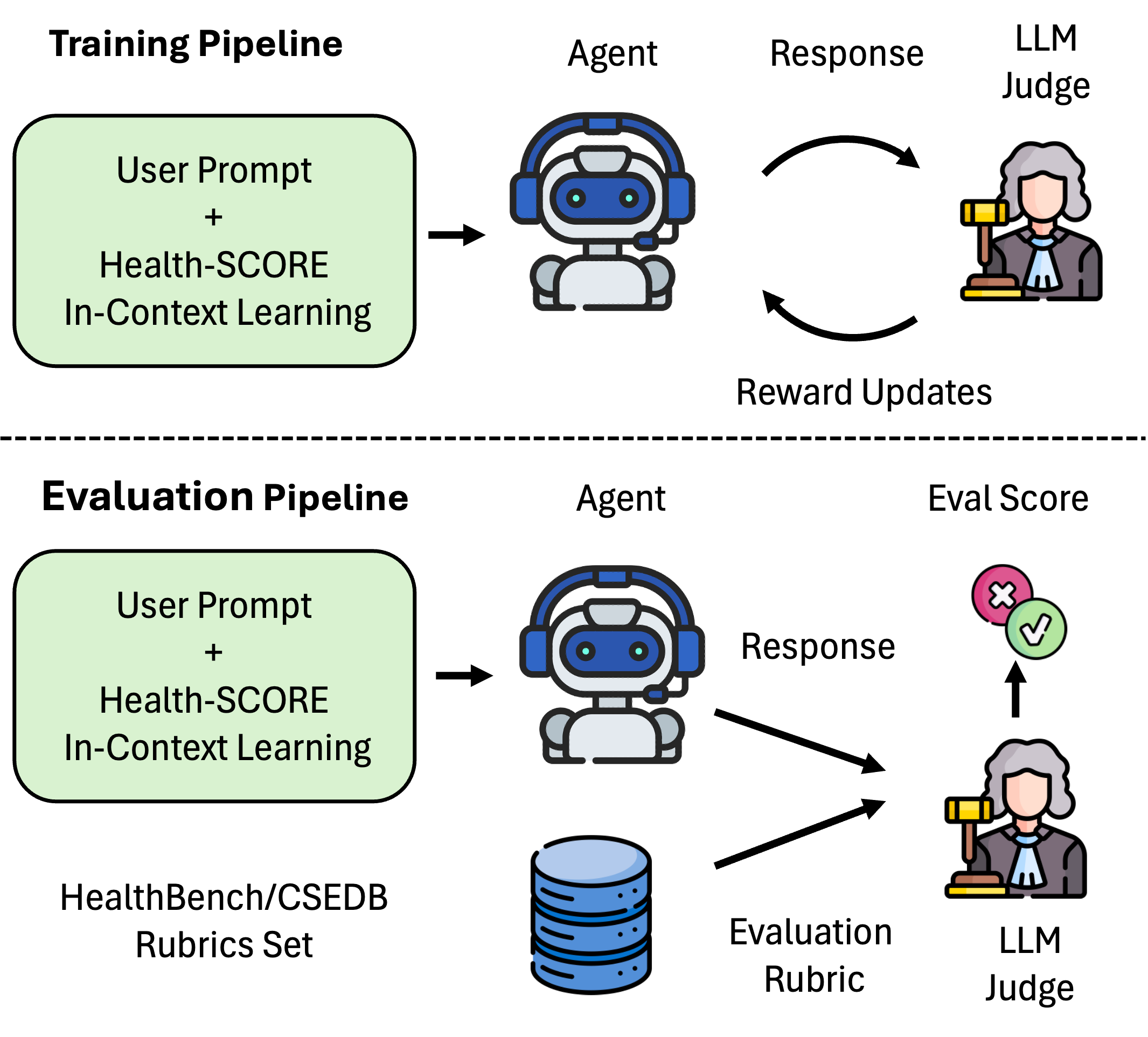}
    \caption{Health-SCORE as the training objective, with evaluation using gold-standard human-authored rubrics.}
    \label{fig:train_eval}
\end{figure}

\section{Experimental Setup}
\subsection{Evaluation Methodology}
\label{sec:Evaluation_Methodology}
We aim to evaluate whether Health-SCORE can (1) function as an effective reward signal for improved model training and (2) support in-context learning for test-time prompt guidance, rather than serving solely as a standalone evaluation framework. To this end, we  designed three sets of experiments (Section \ref{sec:results}) to assess Health-SCORE from these perspectives. Section~\ref{sec:results_rl} examines the benefits of using Health-SCORE as an effective reward for healthcare-specific post-training. Section~\ref{sec:results_icl} examines the benefits of Health-SCORE for in-context learning to generate higher-quality responses at test time, and Section~\ref{sec:results_efficiency} demonstrates its benefits for training efficiency and stability. Furthermore, in order to measure the generalizability of Health-SCORE, we run experiments on both in-domain and out-of-distribution (OOD) data. For in-domain evaluation, we measure performance on held-out conversations from HealthBench–HealthData. These examples match the training split in task categories, data source, and physician-authored rubric construction protocol, but are disjoint at the conversation level. We test generalization using two OOD settings:
\begin{itemize}
\item OOD-Difficulty: HealthBench-Hard, which remains in the same healthcare domain as HealthBench–HealthData but contains unseen and more challenging instances that require deeper reasoning, involve greater ambiguity, and exhibit stricter failure modes.

\item OOD-Dataset: CSEDB \citep{wang2025novel}, which differs from HealthBench not only in data source but also in task formulation and rubric ontology. CSEDB emphasizes safety and effectiveness, rather than the HealthBench evaluation axes. The dataset is originally in Chinese and has been translated into English.
\end{itemize}

\subsection{Evaluation Using Gold-Standard Rubrics}
Figure~\ref{fig:train_eval} shows our evaluation pipeline. We evaluate all models using human-authored, instance-level rubrics from HealthBench and CSEDB, rather than Health-SCORE itself. This design choice avoids circular evaluation and enables a direct assessment of whether models trained or guided by Health-SCORE better align with independent expert judgment. In this framework, Health-SCORE functions as a scalable surrogate supervision signal, while human-authored rubrics serve as the gold standard for evaluation. Following HealthBench, we use GPT-4.1 as the judge for  expert-crafted rubrics. Importantly, the LLM judge does not define the evaluation criteria; it only assesses whether a response adheres to physician-authored criteria.  HealthBench reports large-scale validation of this judge setup on 60,896 meta-examples, finding that GPT-4.1 operates within the natural range of physician inter-rater variability: it matches or exceeds average physician agreement in 5 of 7 themes, ranks in the top half of physicians in 6 of 7 themes, and is never below the bottom third. All reported scores are normalized to the range [0.0,1.0]. Experiment details are available in Appendix~\ref{sec:method_detail}. Furthermore, we conducted experiments to measure the robustness of Health-SCORE to judge selection (Appendix~\ref{sec:judge_bias}).

\subsection{Baselines}
To  measure the effect of our components, we evaluate our method against the following baselines:

\paragraph{Single-Axis} Single-Axis Instruction-only rubrics use a single-axis, generic rubric applied across domains:
\texttt{"You are a helpful assistant. Please generate a response that follows user instructions."}    
This rubric focuses on instruction-following, without domain-specific constraints or multi-dimensional axes. It is lightweight and easy to scale, but lacks task-specific guidance.  

\paragraph{Multi-Axes} Non-Adaptive Multi-Axis rubrics are a set of fixed, multi-axis evaluation criteria covering five axes from HealthBench: communication quality, instruction following, accuracy, context awareness, and completeness. This baseline improves relevance compared to Single-Axis, but is non-adaptive with limited representational power. 

\paragraph{LLM-Generated} Inspired by \citet{gunjal2025rubrics,wang2025infimed}, rubrics are generated by an LLM and specific to each prompt. Given each user prompt, we first ask GPT-4.1 to generate a list of appropriate rubrics in natural language, and then applied those rubrics as reward functions. 

\paragraph{Instance-Specific} HealthBench Instance-specific rubrics are the physician-authored criteria provided in HealthBench, designed specifically for each conversation instance. They yield the most precise and contextually relevant reward signals, but require substantial development effort and domain knowledge, making them costly, difficult to scale, and of limited use for in-context learning.

\begin{figure}[!b] 
    \centering 
    \includegraphics[width=0.5\textwidth]{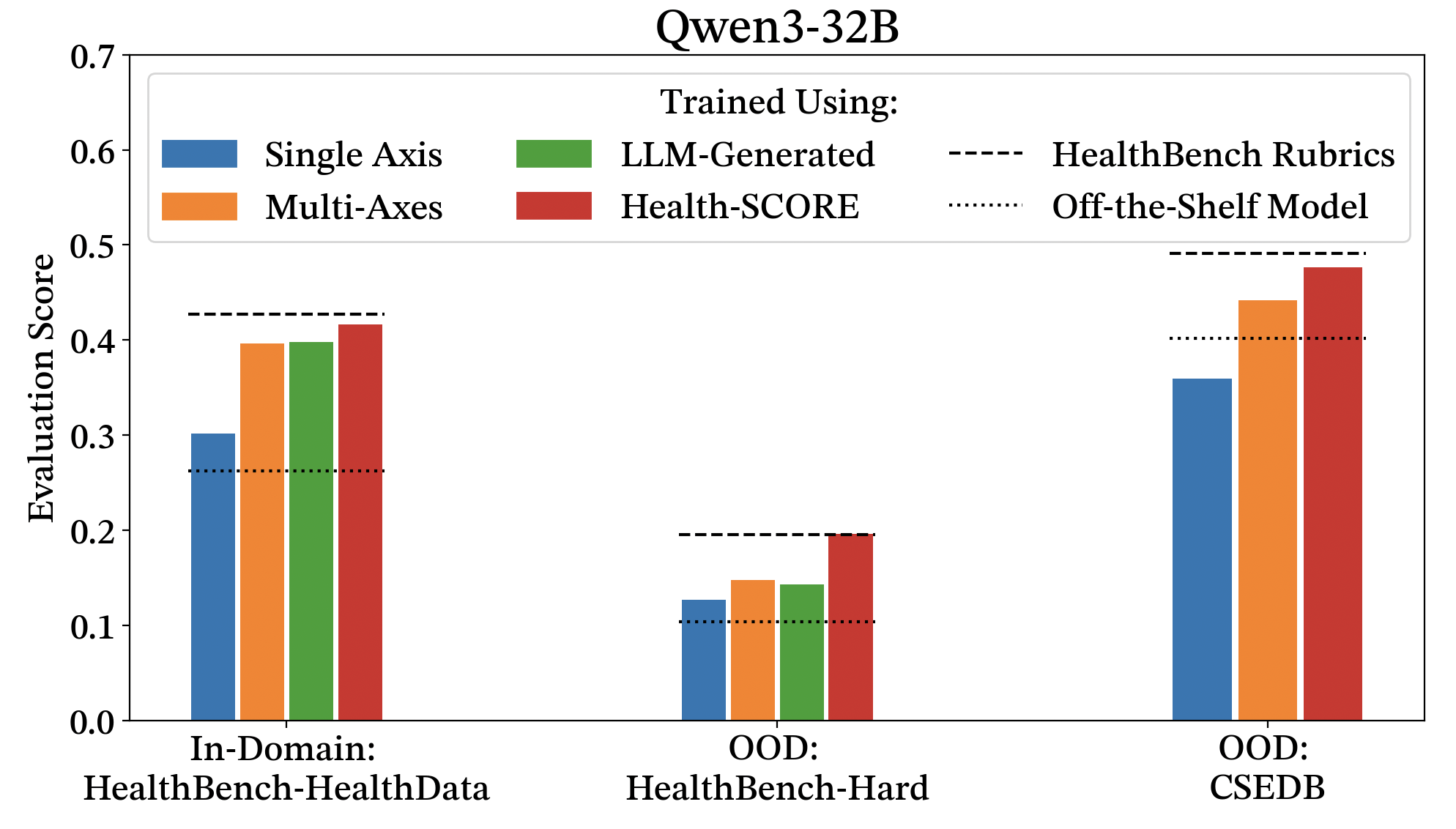}
    \includegraphics[width=0.5\textwidth]{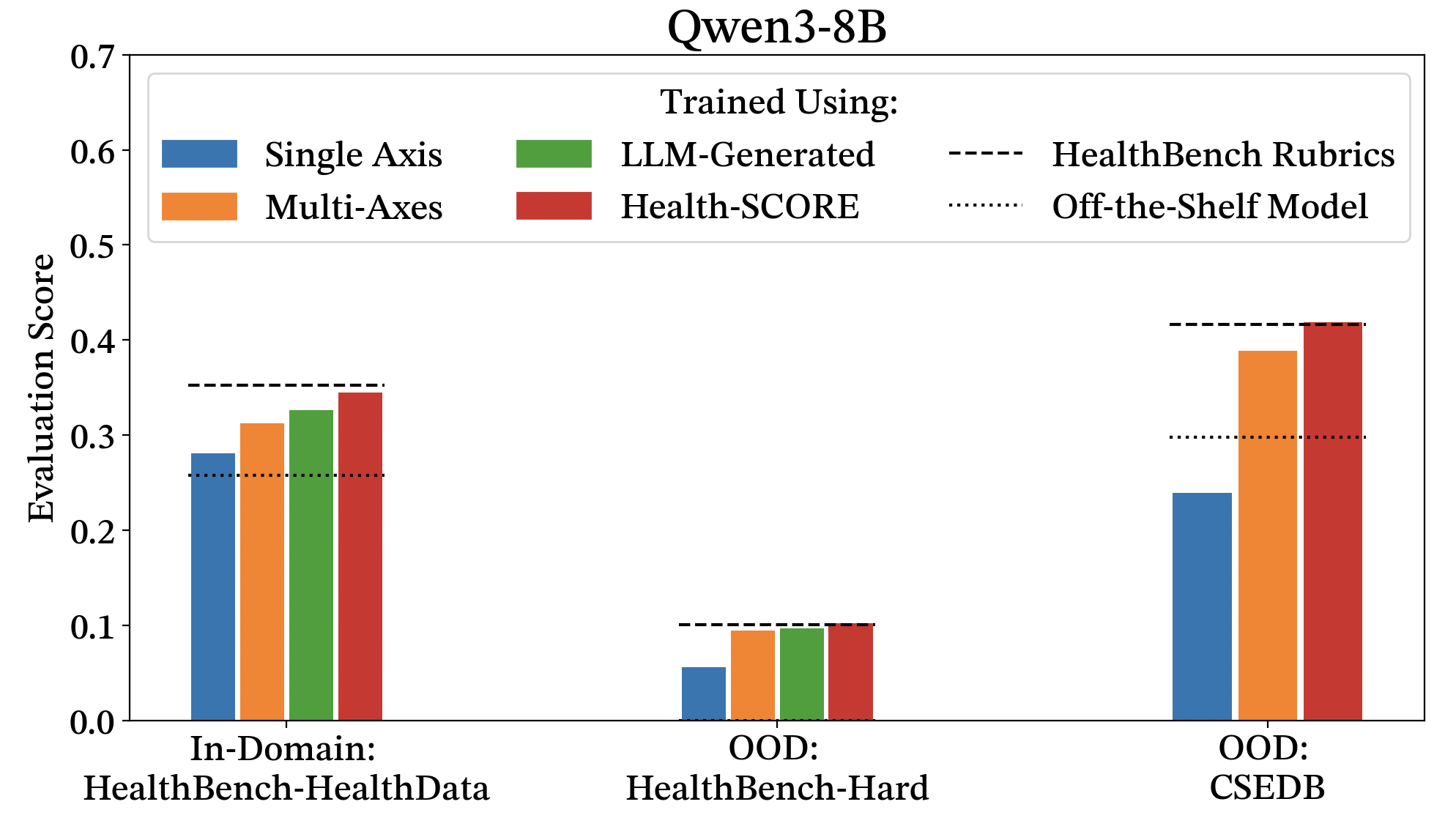}
    \caption{In-Domain and OOD evaluation when models are trained with different rubric types. Dotted/dashed lines correspond to lower/upper bounds.}
    \label{fig:main_result} 
\end{figure}

\begin{figure*}[!t]
  \centering
  \begin{subfigure}{0.32\textwidth}
    \centering
    \includegraphics[width=\linewidth]{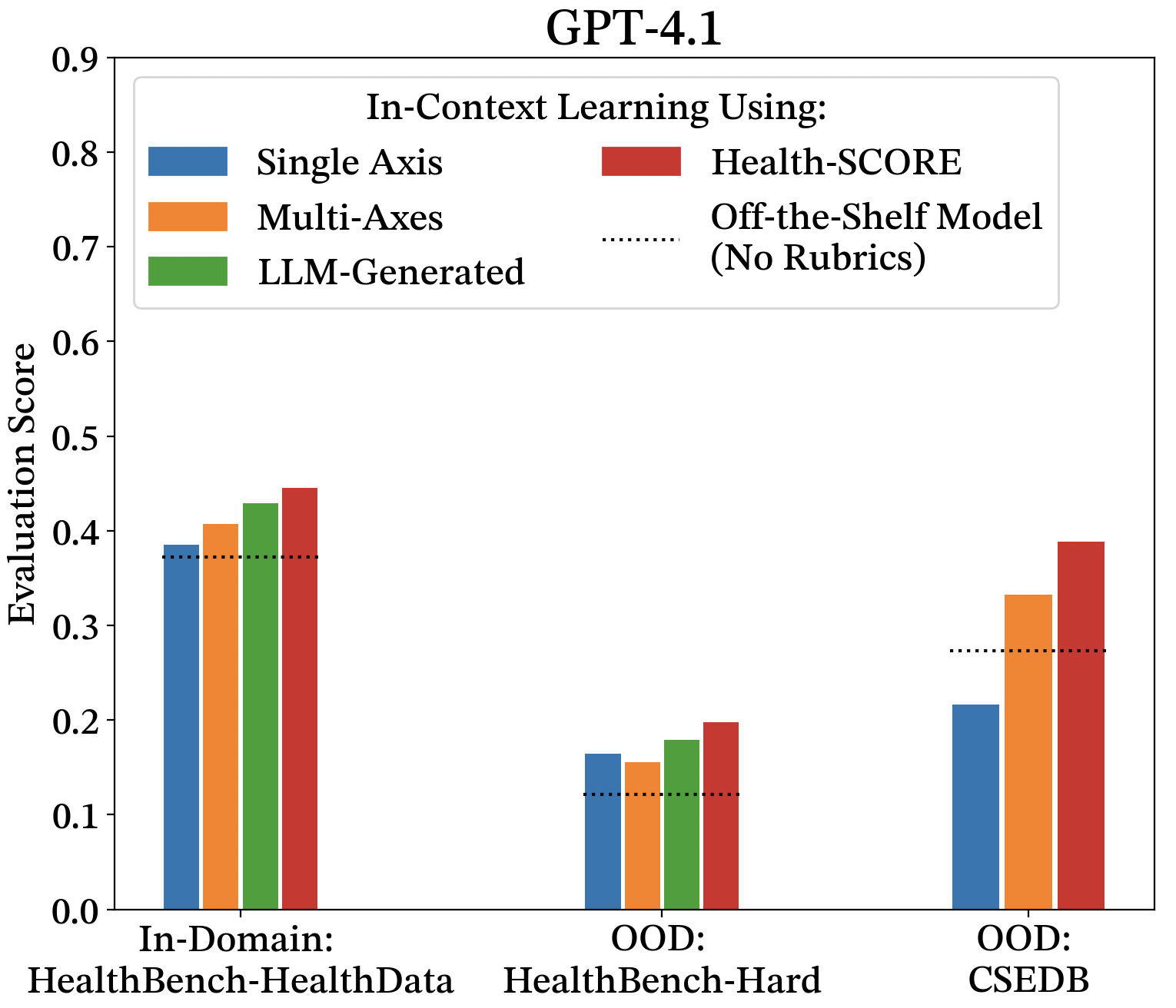}
  \end{subfigure}
  \begin{subfigure}{0.32\textwidth}
    \centering
    \includegraphics[width=\linewidth]{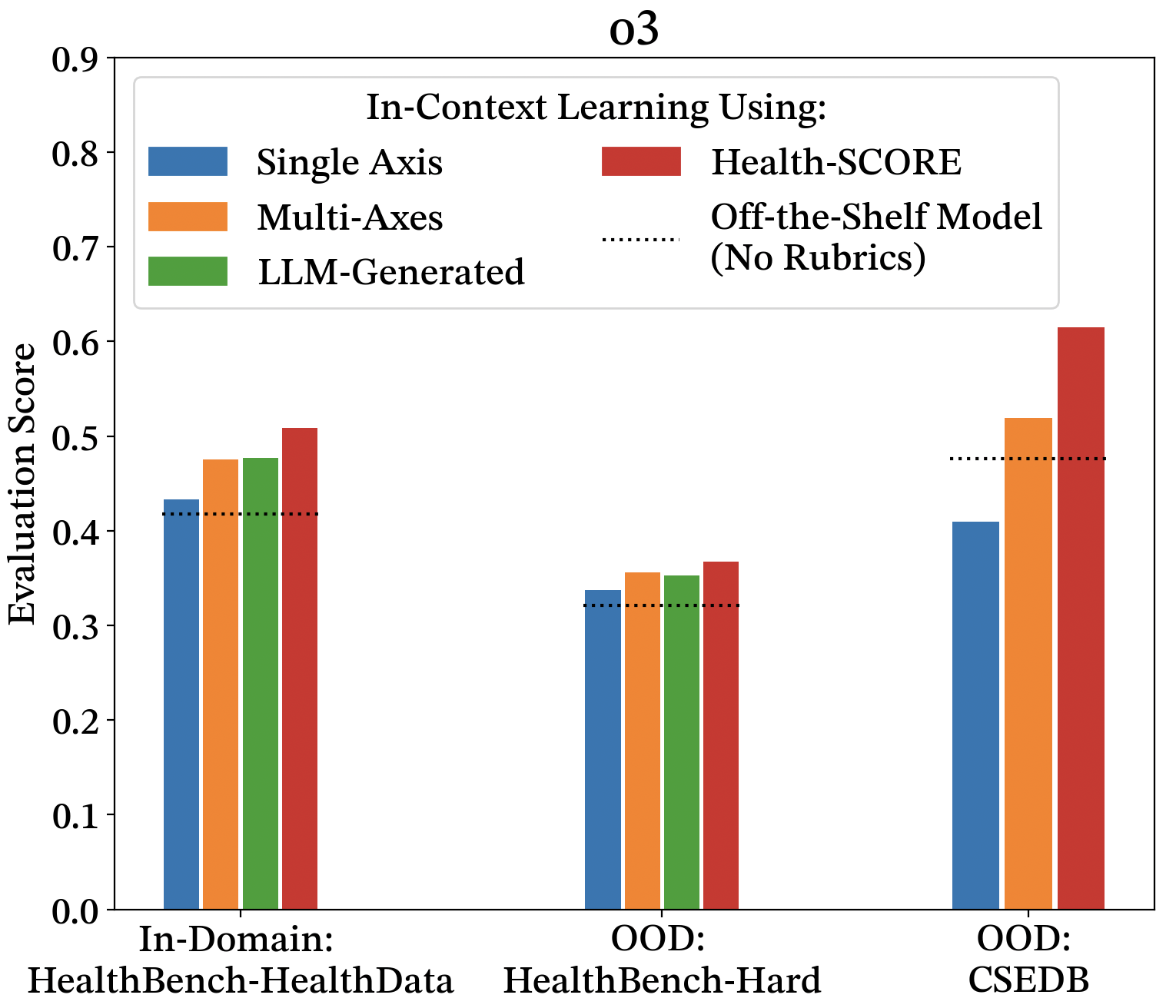}
  \end{subfigure}
  \begin{subfigure}{0.32\textwidth}
    \centering
    \includegraphics[width=\linewidth]{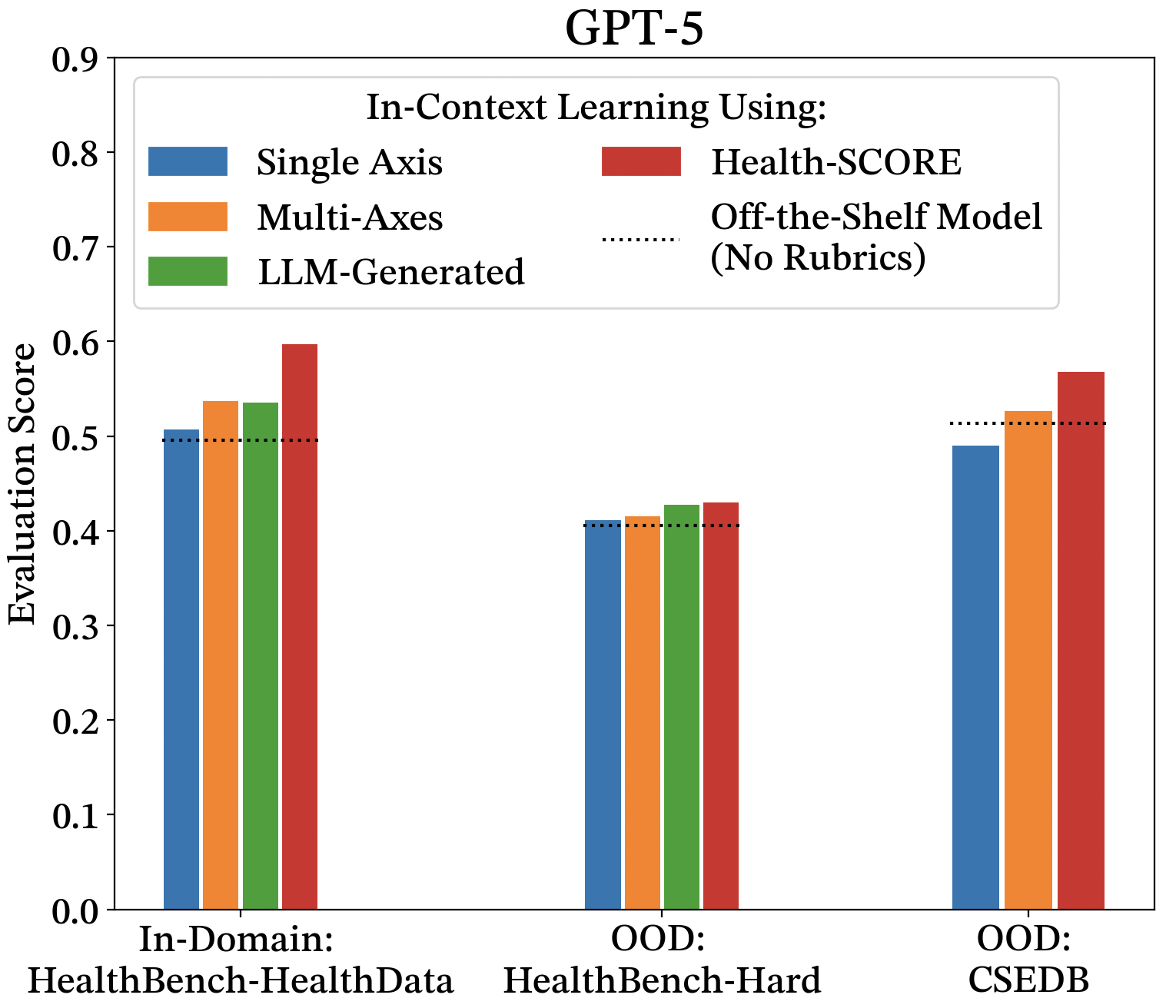}
  \end{subfigure}
  \caption{In-Domain and out-of-distribution evaluation when models are inferred with different rubric types.}
  \label{fig:infer_result}
\end{figure*}

\section{Results}
\label{sec:results}
We evaluate the effectiveness of Health-SCORE from three complementary perspectives: (i) its value as a reward signal during reinforcement learning,  (ii) its utility as test-time guidance via in-context learning without additional fine-tuning and (iii) its effect on training dynamics when included directly in prompts.
First, we show that models trained with Adaptive Health-SCORE achieve stronger alignment with physician-authored evaluation criteria, outperforming baselines in both in-domain and out-of-distribution settings. Second, we demonstrate that using Adaptive Health-SCORE purely at inference time also yields consistent performance gains. Finally, we analyze training dynamics and find that incorporating Health-SCORE into prompts during learning improves sample efficiency and stabilizes policy optimization by steering exploration toward human-relevant criteria. Together, these results establish Adaptive Health-SCORE as a unified mechanism for scalable evaluation, efficient training, and effective test-time improvement of healthcare LLMs.

\subsection{Health-SCORE as RL Reward Signal}
\label{sec:results_rl}

Figure~\ref{fig:main_result} compares the performance of models trained with different rubric reward formulations. All models are evaluated using HealthBench instance-specific rubrics, following the HealthBench evaluation protocol. Across evaluation settings and model sizes, Health-SCORE consistently outperforms alternative reward formulations.
In the in-domain setting, Health-SCORE achieves the strongest performance among reward formulations, surpassing the Single-Axis, Multi-Axis, and Generated Rubrics baselines, while remaining competitive with training on HealthBench instance-specific rubrics. This suggests that adaptively selecting contextually relevant criteria provides a better supervision signal than relying on either fixed rubric dimensions or automatically generated instance-level rubrics.
These gains also extend to OOD settings. On HealthBench-Hard, Health-SCORE shows substantial improvements, particularly for larger models, indicating greater robustness under distribution shift. A similar pattern is observed on CSEDB, which differs from HealthBench in data source, task formulation, and rubric ontology; Health-SCORE achieves the best performance for both model sizes. Taken together, these results show that training with adaptive meta-rubric reward signals improves generalization across domains, datasets, and evaluation settings.

\subsection{In-Context Learning: Better Generation}
\label{sec:results_icl}
In addition to improving training, Health‑SCORE can also be used at inference time to generate higher‑quality responses when incorporated directly into prompts. This addresses one of the key limitations of using predefined instance-based rubrics such as those in HealthBench. Figure \ref{fig:infer_result} shows that Health-SCORE can improve performance even for frontier models such as GPT-5, o3, and GPT-4.1, which were not trained with Health-SCORE, in both in-domain and out-of-distribution settings. This is consistent with prior findings showing that rubric-based prompting can  enhance model performance by making evaluation criteria explicit at generation time \cite{wang2025novel}.

\begin{figure}[tpb]
    \centering
    % \\[-10pt]
    \includegraphics[width=0.45\textwidth]{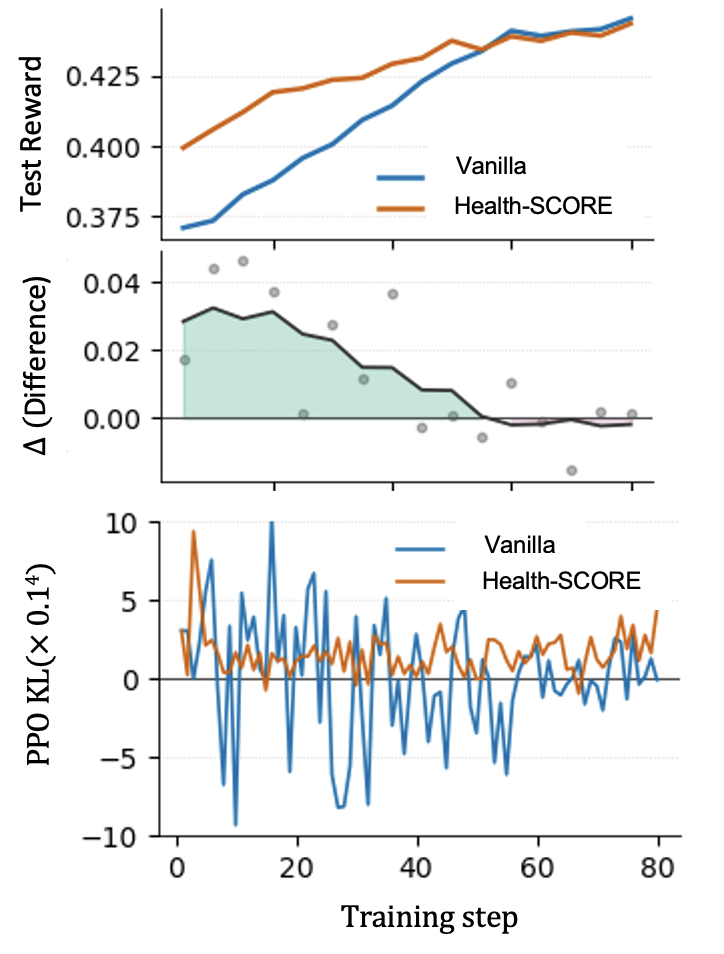}
    \caption{Top: Health-SCORE speeds up RL training convergence, leading to higher $\Delta$ early in training. Bottom: Health-SCORE improves KL training stability.}
    \label{fig:sample_efficiency}
\end{figure}

\subsection{Training Efficiency Improvement}
\label{sec:results_efficiency}
In addition to overall performance gains, we analyze how incorporating Health-SCORE rubrics directly into the prompts  influences the learning dynamics during training. As shown in Figure~\ref{fig:sample_efficiency}, models trained with Health-SCORE prompting consistently achieve higher evaluation scores earlier in training, suggesting that Health-SCORE guides the policy toward desirable behaviors and reduces wasteful exploration. This is further supported by the training stability analysis in Figure~\ref{fig:sample_efficiency}. Health-SCORE prompting results in less volatile policy-update KL compared to training without rubric conditioning, indicating smoother, less noisy, and more controlled policy updates. These results suggest that Health-SCORE is beneficial not only as a reinforcement learning reward signal, but also as an explicit conditioning mechanism during generation. By shaping the exploration space, Health-SCORE enables faster and more stable convergence without compromising final performance.

\subsection{Ablation: Adaptive Selection Mechanism}
\label{sec:Ablation_Adaptive}

To evaluate the impact of the adaptive selection mechanism, we further conduct ablation studies. Table~\ref{tab:ablation} reports the results of these experiments. As shown, the adaptive selection mechanism consistently improves evaluation scores across all settings and models. This holds true both when the mechanism is applied during training (as in HealthBench:HealthData) and when it is used only at inference time through in-context learning. The effect is particularly strong for weaker models, e.g. Qwen3-8B. In these cases,  including all rubric criteria indiscriminately can dilute the learning/guidance signal by introducing many constraints, some of which may not be relevant to the specific conversation. Adaptive selection mitigates this issue by focusing on the most pertinent criteria.

\begin{table}[t]
\renewcommand{\arraystretch}{1.2}
\centering
\scriptsize
\begin{tabular}{|c|c|c|c|}
\hline
\textbf{Setup} & \textbf{Model} & \textbf{Non-Adaptive} & \textbf{Adaptive}\\
\hline
\multirow{5}{*}{\makecell[c]{\textbf{HealthBench:}\\\textbf{Health Data}}}
& Qwen3-8B   & 0.051 & 0.345 \\
& Qwen3-32B  & 0.279 & 0.416 \\ 
& GPT-4.1    & 0.328 & 0.445 \\
& o3         & 0.391 & 0.509 \\
& GPT-5      & 0.486 & 0.597 \\
\hline
\multirow{5}{*}{\makecell[c]{\textbf{HealthBench:}\\\textbf{Hard (OOD)}}}
& Qwen3-8B   & 0.000 & 0.102 \\
& Qwen3-32B  & 0.073  & 0.196 \\
& GPT-4.1    & 0.136  & 0.198 \\
& o3        & 0.291  & 0.368 \\
& GPT-5      & 0.397  & 0.429 \\
\hline
\multirow{5}{*}{\makecell[c]{\textbf{CSEDB (OOD)}}}
& Qwen3-8B   & 0.263 & 0.418 \\
& Qwen3-32B  & 0.393 & 0.476 \\
& GPT-4.1    & 0.244 & 0.388 \\
& o3         & 0.445 & 0.615 \\
& GPT-5      & 0.491 & 0.568 \\
\hline
\end{tabular}
\caption{Selection mechanism ablation study.}
\label{tab:ablation}
\end{table}

\section{Per-Axis Performance Analysis} \label{sec:axis_analysis}
\begin{figure}[ht] 
    \centering
    \includegraphics[width=0.49\textwidth]{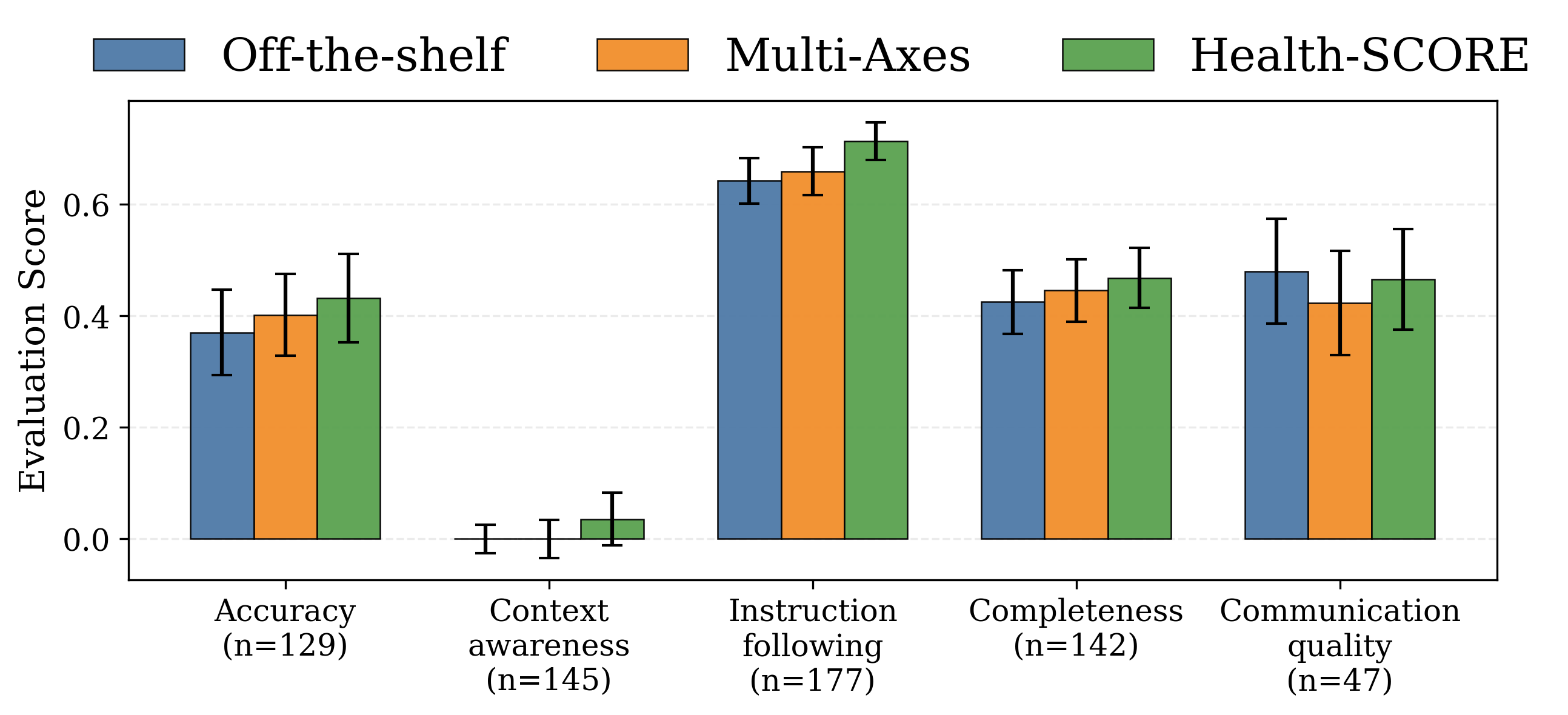}
    \caption{Qwen3-32B performance on HealthBench Health data task, along different dimensions.}
    \label{fig:qwen32b_healthdata}
\end{figure}

In addition to overall benchmark scores, we also examine model performance across individual HealthBench evaluation axes. Figure \ref{fig:qwen32b_healthdata} demonstrates that Health-SCORE improves model performance along most expert-defined rubric dimensions, including accuracy, instruction following, and completeness. Notably, gains are not uniform across all axes. Communication quality exhibits similar performance across methods. This axis-level analysis provides further evidence that Health-SCORE retains the structure of human evaluation criteria and that adaptive rubric selection enables more targeted improvements than either fixed task-level rubrics or generic instruction-following objectives.

\section{Conclusion}

Rubric-based evaluation is essential for assessing open-ended LLM outputs in safety-critical domains such as healthcare. However, designing high-quality rubrics is time-consuming and resource-intensive, which limits the scalability of rubric-based evaluation in real-world settings. In this work, we introduce Health-SCORE, a scalable and generalizable framework that reduces rubric development effort while maintaining strong evaluation performance. Through extensive experiments, we show that Health-SCORE serves as an effective alternative to human-authored rubrics, both as an inference-time guidance mechanism and as a structured reward signal for reinforcement learning. Models trained or guided using Health-SCORE achieve improved performance and training efficiency when evaluated against expert instance-level criteria. These results highlight Health-SCORE’s potential as a practical and scalable solution for evaluating and optimizing healthcare LLMs.

\section{Limitations}  
  
This work makes several assumptions and has limitations that should be considered when interpreting the results.  Health-SCORE relies on LLMs for multiple stages of the pipeline, including rubric embedding, relevance-based selection, and rubric evaluation during reward computation. While LLM-as-a-judge approaches have been shown to correlate well with human judgments in prior work, automated evaluation may still exhibit biases or inconsistencies, particularly in safety-critical domains such as healthcare. Although our final evaluation is conducted using human expert–authored, instance-level rubrics, we do not fully eliminate reliance on automated judges during training. Future work could further improve this by incorporating human-in-the-loop validation or multi-judge consensus mechanisms to improve robustness.

Our approach assumes that human-authored rubrics encode reusable, higher-level evaluation criteria that can be meaningfully abstracted across instances. The clustering-based Health-SCORE rubric construction process, while effective in practice, involves heuristic choices as well as manual refinement and outlier detection step to refine clusters and remove case-specific or noisy rubric items. These design decisions may introduce some level of minor subjectivity. We try to minimize this by validating the resulting Health-SCORE rubrics through downstream performance and axis-level analysis, but alternative abstraction strategies may also yield different Health-SCORE rubric representations.

Although we used the original ground-truth HealthBench weights for evaluation, during training, all selected rubric items contributed equally in our reward formulation, and rubric satisfaction was treated as a discrete outcome. This simplifies reward computation and stabilizes training but does not capture differences in criterion severity or importance that may be reflected in human evaluation.  More expressive reward formulations that incorporate graded satisfaction or learned rubric weights are a promising direction for future work.

We focus our experiments on the health data tasks subset of HealthBench, motivated by its complexity and relevance to structured clinical reasoning. Although we evaluate generalization on HealthBench-Hard and CSEDB, our findings may not imply universal applicability across all healthcare tasks or non-medical domains. Extending Health-SCORE to broader task families and validating them under substantially different evaluation ontologies remain important future directions.

A potential risk of rubric-based reinforcement learning is reward hacking, where models learn to exploit rubric-visible features without improving the underlying quality of their responses. For example, models may adopt templated phrasing, add excessive disclaimers, or overemphasize criteria that are easy for the judge to detect while neglecting less explicit but clinically important aspects of the task. Although our experiments show that Health-SCORE does not increase response length—and in some settings reduces it—shorter responses do not rule out these forms of shortcut behavior. Mitigating this risk requires evaluation with independent human-authored criteria, qualitative inspection for repetitive or templated behavior, diverse and adversarial test cases, and periodic updates to rubric and judge design to reduce exploitable regularities.

\bibliography{custom}

\clearpage

\appendix 

\begin{table*}[t!]
\renewcommand{\arraystretch}{1.3}
\setlength{\tabcolsep}{6pt}
\centering
\small
\begin{tabular}{|c|c|c|c|c|c|c|}
\hline
{Setting} &
{Judge} &
{Accuracy} &
\makecell[c]{{Context}\\[3pt]{Awareness}} &
\makecell[c]{{Instruction}\\[3pt]{Following}} &
{Completeness} &
{Communication} \\
\hline
\multirow{2}{*}{\makecell[c]{{Off-the-shelf}}}
& GPT-4.1 & 0.6438 & 0.2251 & 0.7373 & 0.5726 & 0.5418 \\
& o3      & 0.6501 & 0.2233 & 0.7464 & 0.5652 & 0.5423 \\
\hline
\multirow{2}{*}{\makecell[c]{{HealthSCORE}}}
& GPT-4.1 & 0.6861 & 0.4968 & 0.7759 & 0.5929 & 0.6519 \\
& o3      & 0.6878 & 0.4907 & 0.7808 & 0.5907 & 0.6636 \\
\hline
\end{tabular}
\caption{Health-SCORE cross-judge consistency.}
\label{tab:crossjudge-results}
\end{table*}

\section{Potential Risks}  

Our work introduces adaptive Health-SCORE to enable scalable rubric-based evaluation and training of large language models, particularly for healthcare tasks. A primary risk is overgeneralization, where abstracted rubrics may be applied beyond contexts that require instance-specific or domain-expert judgment, potentially leading to misplaced trust in model outputs. Because Health-SCORE is derived from existing human-authored criteria, its rubrics may also propagate or amplify implicit biases present in the original benchmarks, which could disproportionately affect underrepresented populations if deployed without auditing. Additionally, when used as reward functions for reinforcement learning, rubric-based supervision may incentivize models to optimize for measured criteria while neglecting unmeasured but clinically relevant factors. We emphasize that Health-SCORE is intended to complement rather than replace human evaluation, and our experiments explicitly rely on independent, expert-authored instance-level rubrics for final assessment. To mitigate these risks, we recommend human-in-the-loop oversight, periodic bias audits of abstracted criteria, careful scoping of applicability, and efficiency-aware implementation to limit unnecessary computational and environmental costs.

\section{Rubrics Statistics} \label{sec:rubric_stat}

Table \ref{tab:rubrics_tokens} summarizes the structural properties of different types of rubrics used in our experiments. We report three statistics, each averaged over all conversations in the evaluation set. Number of rubrics denotes the average count of individual rubric criteria applied to a single conversation. For non-adaptive methods, such as Single-axis, Multi-axes, and Health-SCORE (Non-adaptive), this value is fixed by design and identical across examples. For adaptive methods, such as LLM-Generated, Instance-specific, and Health-SCORE (Adaptive), the number varies per prompt and is computed by counting only the rubric items used for that conversation, then averaging across the dataset. Number of tokens measures the average length of the rubric text for each conversation. The adaptive column indicates whether the rubric set varies across conversations. To assess whether our meta-rubric-based rewards lead to reward overoptimization, such as encouraging unnecessarily verbose responses, we conducted an additional quantitative analysis of response length. We find that Health-SCORE does not increase response length: the average final answer length is 363.88 tokens, compared with 683.22 tokens for the original model, 601.39 for the fixed-axis baseline, and 424.82 for the human-rubric reward setting. Reasoning length follows a similar pattern, with Health-SCORE averaging 422.38 tokens, compared with 536.28 for the original model, 503.39 for the fixed-axis baseline, and 447.19 for the human-rubric setting. Although our experiments show that Health-SCORE does not increase response length—and in some settings reduces it—shorter responses do not rule out these forms of shortcut behavior. Mitigating this risk requires evaluation with independent human-authored criteria, qualitative inspection for repetitive or templated behavior, diverse and adversarial test cases, and periodic updates to rubric and judge design to reduce exploitable regularities.

 \begin{table*}[ht]  
\centering  
\begin{tabular}{lrrr}  
\hline  
Rubric Method            & Number of Rubrics & Number of Tokens & Adaptive\\  
\hline  
Single-Axis            & 1              & 15   & No         \\  
Multi-Axes             & 5              & 158  & No         \\  
LLM-Generated         & 7.8           & 242.5    & Yes    \\  
Instance-Specific & 10.5          & 452.7    & Yes    \\  
Health-SCORE (Non-Adaptive)      & 29             & 1117 & No      \\  
Health-SCORE (Adaptive)       & 11.5          & 431.4 & Yes     \\  
\hline  
\end{tabular}  
\caption{Rubric and token statistics for different methods. Numbers are averaged over conversations.}  
\label{tab:rubrics_tokens}  
\end{table*}

\section{ LLM-as-a-Judge Robustness Analysis} \label{sec:judge_bias}
\begin{figure}[ht] 
    \centering
    \includegraphics[width=0.49\textwidth]{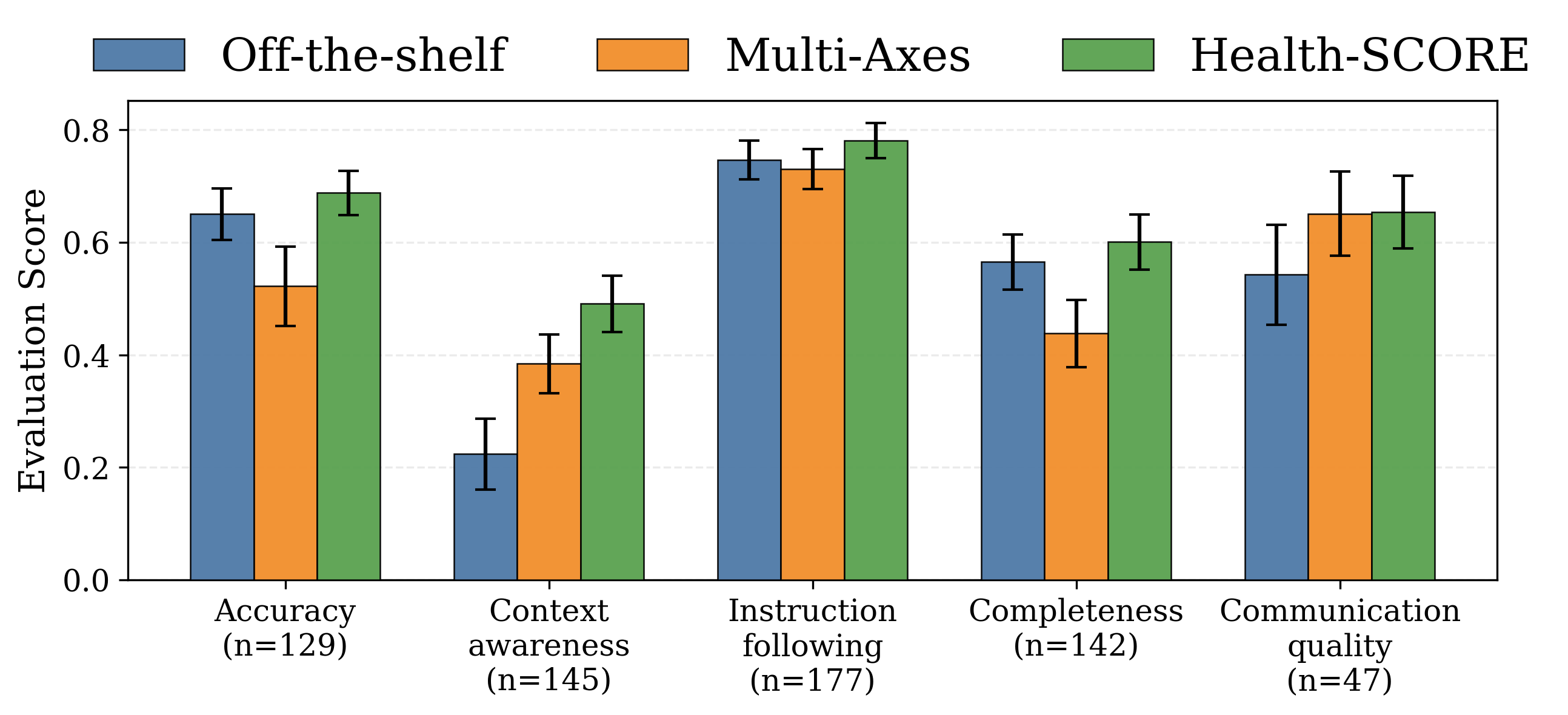}
    \caption{GPT-5 performance on Healthbench Healthdata task, when the judge is o3.}
    \label{fig:gpt5_healthdata_o3judge}
\end{figure}

\begin{figure}[ht] 
    \centering
    \includegraphics[width=0.49\textwidth]{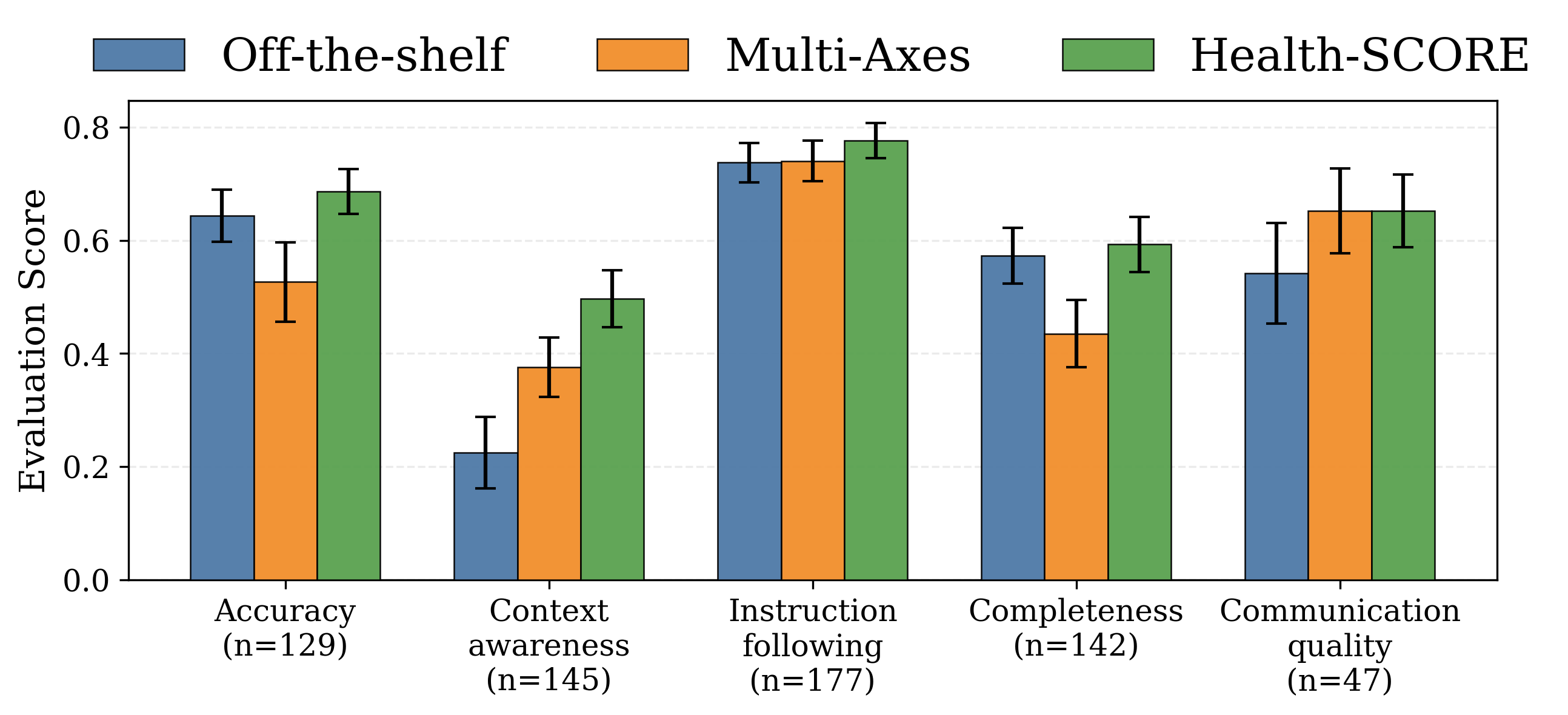}
    \caption{GPT-5 performance on HealthBench Health data task. Bars show mean scores for three evaluation strategies: Off-the-shelf Baseline (blue), Multi-Axes (orange), and Health-SCORE (green). Dimensions include Accuracy (n=129), Context Awareness (n=145), Instruction Following (n=177), Completeness (n=142), and Communication Quality (n=47).}
    \label{fig:gpt5_healthdata}
\end{figure}

We tested whether our evaluation conclusions are robust to the choice of LLM judge by re-running the full HealthBench evaluation with a second judge. Evaluation examples (instruction/query, context, and a single frozen candidate response from GPT-5) and inference systems (Off-the-shelf, Multi-Axes, and Health-SCORE) are identical to Section~\ref{sec:axis_analysis}. Responses were judged by either GPT-4.1 (Figure~\ref{fig:gpt5_healthdata}) or o3 (Figure~\ref{fig:gpt5_healthdata_o3judge}) using the same HealthBench rubric across five axes (accuracy, context awareness, instruction following, completeness, communication).

The o3 judge reproduced the same qualitative conclusions across axes. As shown in Figure ~\ref{tab:crossjudge-results}, there is high cross-judge consistency. Under GPT-4.1 and o3 respectively, Health-SCORE improves over the off-the-shelf baseline on accuracy (0.6438 to 0.6861; 0.6501 to 0.6878), context awareness (0.2251 to 0.4968; 0.2233 to 0.4907), Instruction Following (0.7373 to 0.7759; 0.7464 to 0.7808), completeness (0.5726 to 0.5929; 0.5652 to 0.5907), and communication (0.5418 to 0.6519; 0.5423 to 0.6636). The closely matched improvement margins across judges indicate that the observed gains are not dependent on a single GPT-4.1 judge and are less likely to reflect judge-specific preference amplification.

\section{Implementation Details} \label{sec:method_detail}

To implement the  GRPO algorithm  policy update, we construct prompts by placing the rubric content in the system message, followed by the task-specific user prompt, at each training step. 
For each prompt, eight rollouts are generated. To make optimization with large models feasible, we micro-batch the rollouts together with the reference model’s log-probabilities. The reference model provides log-probs for KL computation and helps stabilize updates. We apply an adaptive KL controller with a low-variance KL penalty (kl\_coef = 1e-4) and a target KL of 0.001 to keep the fine-tuned policy close to the reference. 
Finally, the GRPO estimator uses the judge-provided rewards and reference log-probs to compute advantages and perform mini-batch policy updates over multiple epochs. To speed up reward computation, we invoke the judge asynchronously with a concurrency throttle, allowing many <prompt, response, rubric> triplets to be evaluated in parallel without overloading the judge. 

Following the HealthBench setup, we ensure that each rubric is applied to the final decision, following the last </think> token, rather than any intermediate chain-of-thought content. Rewards are computed with conversation context in mind: model input is a list (system + user messages) and the judge receives rubric text alongside the full prompt and final response, enabling it to assess system-prompt steerability and multi-turn consistency. This enforces global constraints (tone, persona, forbidden content) rather than treating each model turn as an independent checklist.

Model comparisons are conducted using the same evaluation protocol and metrics across all baselines. 
% Where reinforcement learning is involved, each configuration is trained once due to computational constraints; however, 
Uncertainty in evaluation is captured through bootstrap resampling of test examples. Bootstrap standard deviation is computed by resampling with replacement 1000 times, computing the mean for each resample, then taking the standard deviation of those bootstrap means. 

All post-training experiments were conducted on a single node equipped with 8 NVIDIA A100 GPUs. The total GPU usage for training and evaluation was approximately 30 GPU-hours.
The HealthBench - HealthDatad Tasks dataset comprises 477 examples and HealthBench Hard  includes 1,000  examples.

\section{Ablation: Adaptive Selection Mechanism Robustness} \label{sec:app_adaptive_select}

\begin{figure*}[p]
    \centering
    \includegraphics[width=0.95\textwidth,height=0.78\textheight,keepaspectratio]{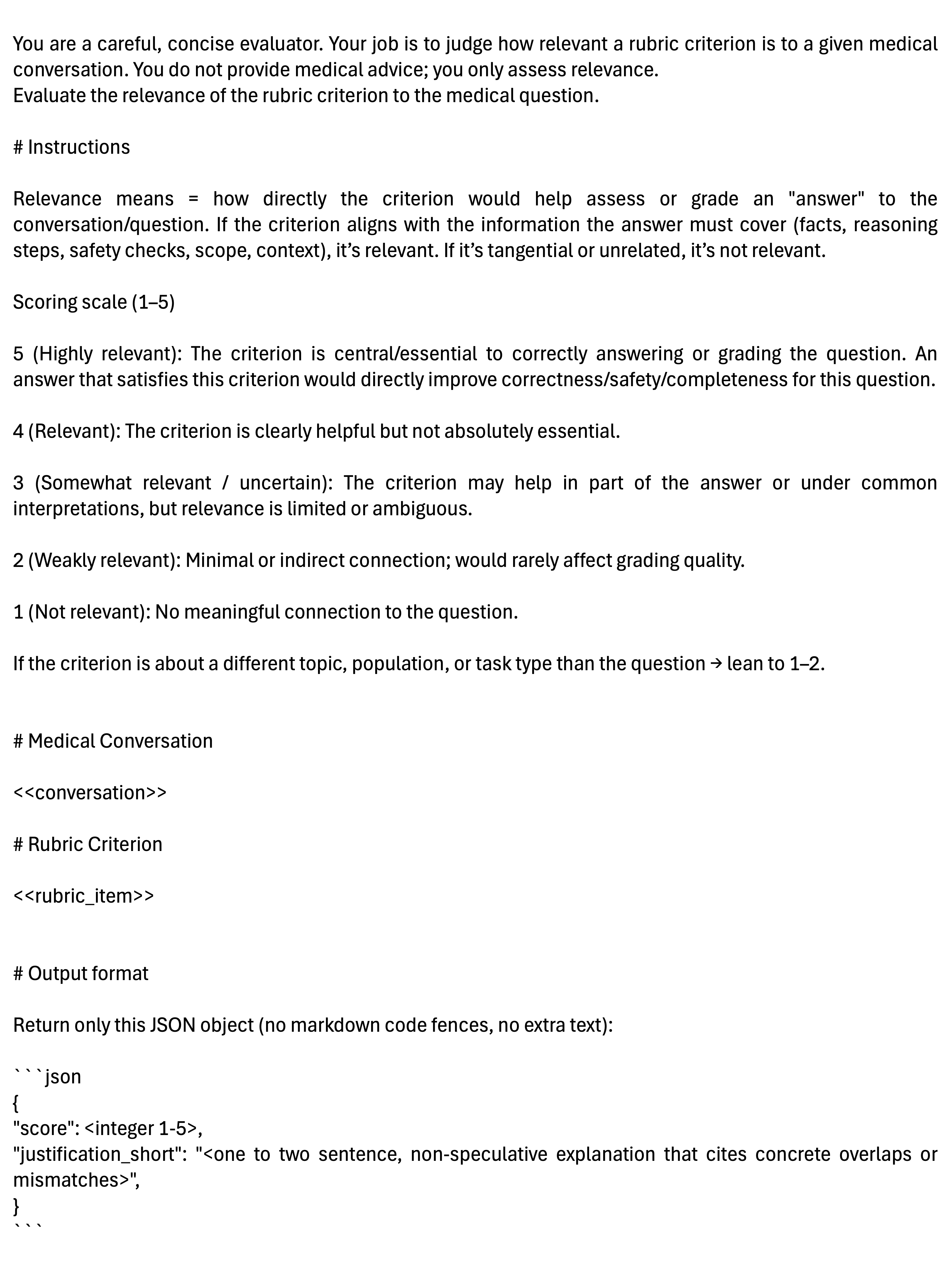}
    \caption{Prompt used for adaptive Health-SCORE criterion selection.}
    \label{fig:selection_prompt_appendix}
\end{figure*}

We used the prompt in Figure \ref{fig:selection_prompt_appendix} for the adaptive criteria-selection mechanism. The selector scores each candidate Health-SCORE criterion from 1 to 5 according to its relevance to the input conversation and retains criteria with scores above the fixed threshold. This threshold was chosen once during the Health-SCORE design process using only the initial HealthBench HealthData rubrics. It was not modified for different policy models, prompting models, datasets, or out-of-distribution evaluations.

To assess selector reliability, we measured agreement between the adaptive selector and the original human-annotated HealthBench rubrics, treating the human-authored rubric assignments as the reference set of applicable criteria. We evaluated agreement across four selector models (GPT-4.1, GPT-4.1-mini, GPT o4-mini, and GPT-5.2), three reasoning levels (low, medium, and high), temperatures ranging from 0.2 to 2, and three prompt variants. Across these settings, the selector achieved 83\% alignment accuracy with the human-annotated rubric assignments, with a standard deviation of 1\%. This low variation across models, temperatures, reasoning levels, and prompt settings suggests that the adaptive selection mechanism is stable under common sources of selector variation. In addition, all reported final evaluations use independent, human-annotated instance-level rubrics rather than the adaptive selector. Therefore, improvements measured on the evaluation sets do not depend on the selector used to construct training rewards or in-context criteria.

\section{Ablation: Retrieval Baseline} \label{sec:retrieval_baseline}

Furthermore, to compare Health-SCORE against a purely retrieval-based baseline, we evaluate retrieval performance on HealthBench using Qwen-8B under the same evaluation protocol.

These results provided in Table ~\ref{tab:healthscore_comparison} indicate that Health-SCORE provides additional gains beyond simply retrieving relevant rubric information.

\begin{table}[b]
\centering

\begin{tabular}{|l|c|}
\hline
Method & Score \\
\hline
Single-axis baseline & 0.2779 \\
Retrieval baseline & 0.3070 \\
LLM-generated  & 0.3206 \\
Health-SCORE & 0.3447 \\
\hline
\end{tabular}
\caption{Performance comparison on of Qwen-8B using HealthBench- HealthData.}
\label{tab:healthscore_comparison}
\end{table}

\section{Health-SCORE Criteria List}  \label{sec:criteria_list}

\begin{figure*}[p]
    \centering
    \includegraphics[width=\textwidth,height=0.88\textheight,keepaspectratio]{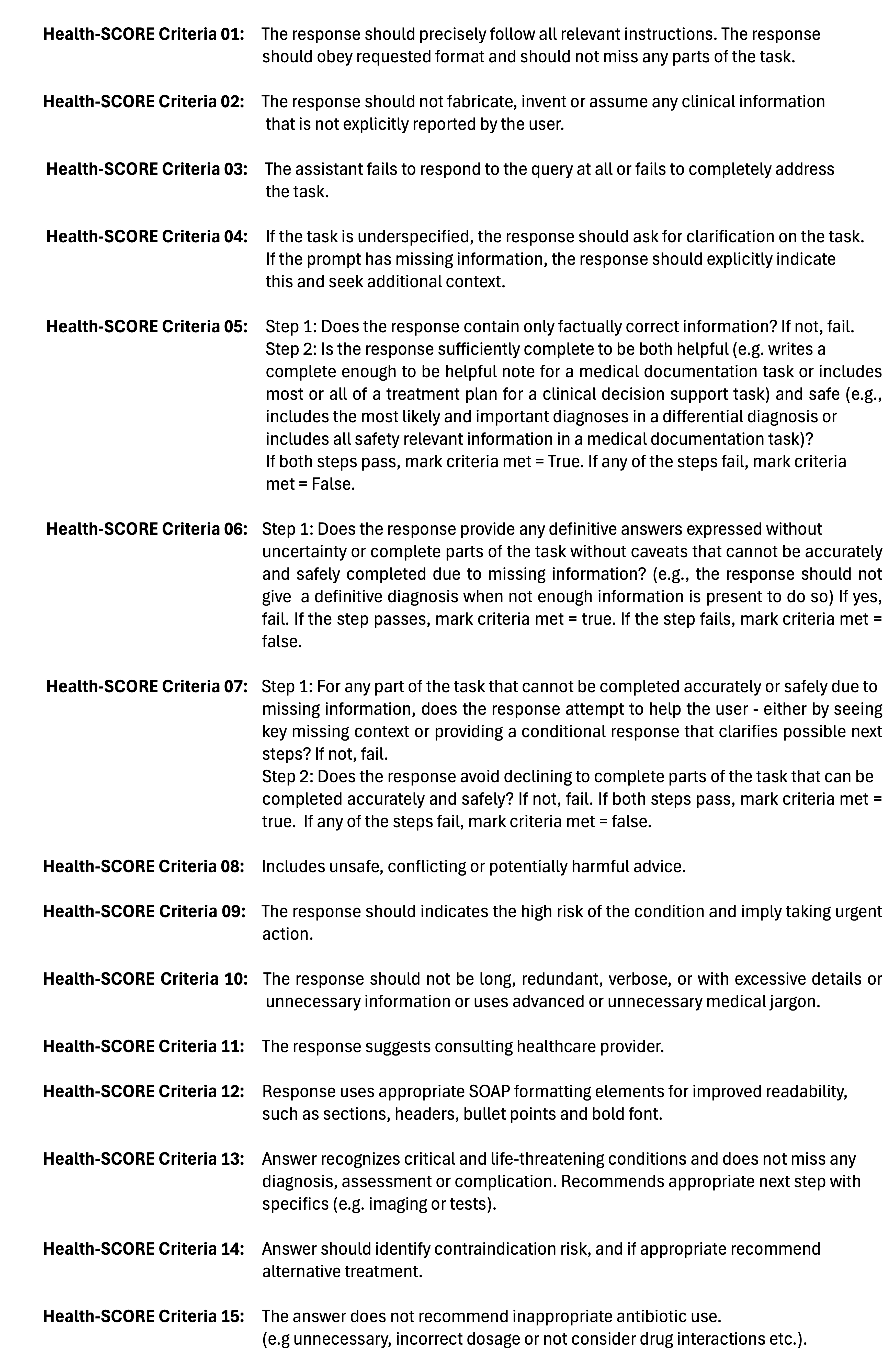}
    \caption{Health-SCORE criteria list, part 1.}
    \label{fig:hs_rubric_list_1}
\end{figure*}

\begin{figure*}[p]
    \centering
    \includegraphics[width=\textwidth,height=0.68\textheight,keepaspectratio]{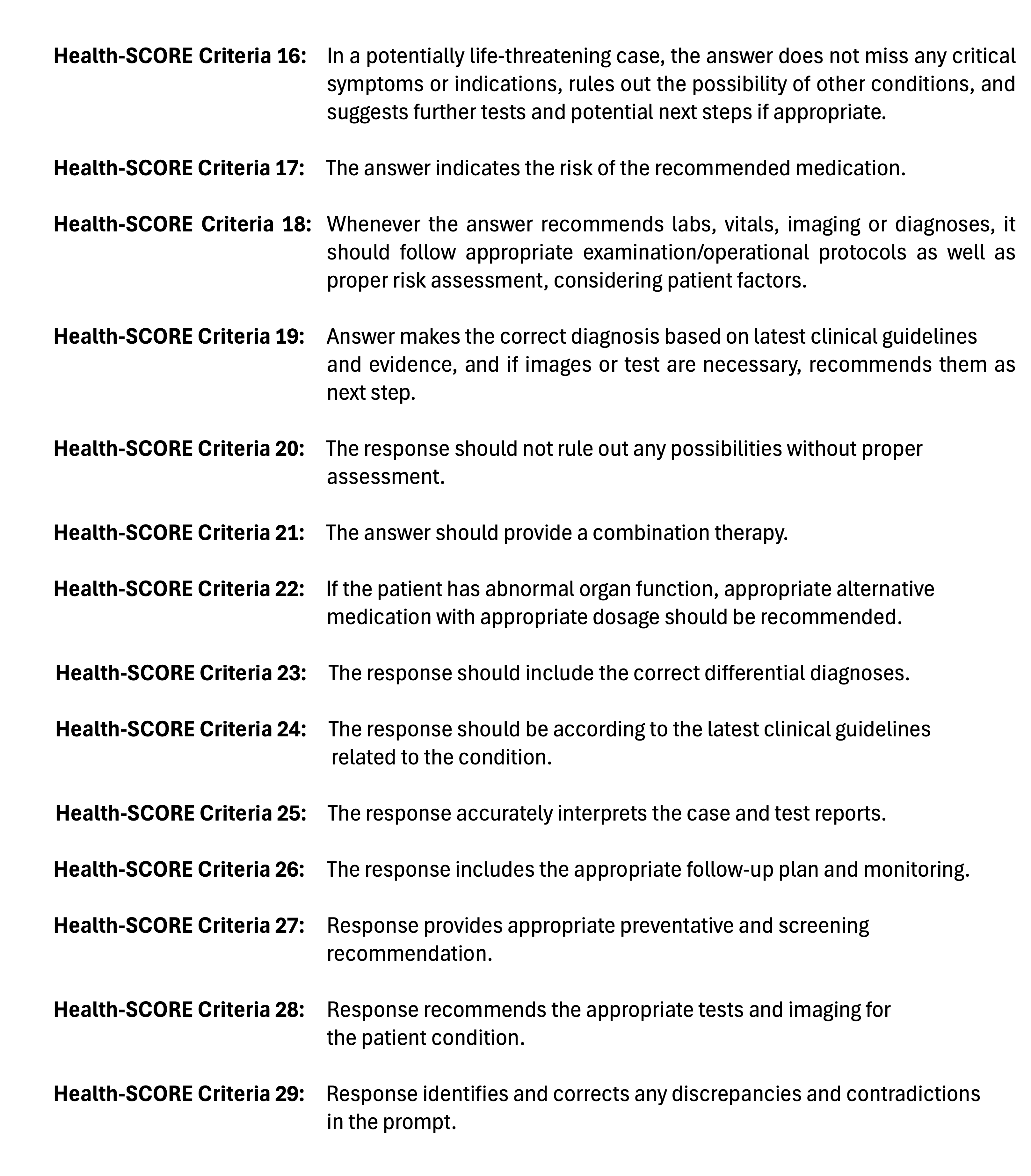}
    \caption{Health-SCORE criteria list, part 2.}
    \label{fig:hs_rubric_list_2}
\end{figure*}

Figures~\ref{fig:hs_rubric_list_1} and~\ref{fig:hs_rubric_list_2} provide the complete set of Health-SCORE criteria used in our experiments.

\section{Health-SCORE Meta-criteria Categorization} 

Table~\ref{tab:healthscore_categories} provides a categorization of the 29 Health-SCORE criteria into seven higher-level categories based on the type of quality or safety aspect they evaluate. These categories are: (1) Instruction Adherence and Response Quality, covering task completion, formatting, clarity, and appropriate handling of missing information; (2) Information Grounding and Internal Consistency, capturing hallucination avoidance, faithful interpretation of provided information, and contradiction detection; (3) Patient Safety, Urgency, and Critical Conditions, focusing on harmful advice, recognition of high-risk presentations, and appropriate escalation; (4) Diagnosis, Differential Diagnosis, and Evidence, assessing diagnostic correctness, differential coverage, and adherence to clinical guidelines; (5) Testing, Imaging, and Clinical Assessment, evaluating the appropriateness of recommended examinations, laboratory tests, and imaging; (6) Treatment and Medication Management, covering contraindications, medication safety, antibiotic use, treatment selection, and dose adjustment; and (7) Follow-up, Referral, and Preventive Care, assessing recommendations for monitoring, clinician follow-up, screening, and prevention.

\section{Qualitative Example}
\begin{figure*}[p]
    \centering
\includegraphics[width=0.95\textwidth,height=0.78\textheight,keepaspectratio]{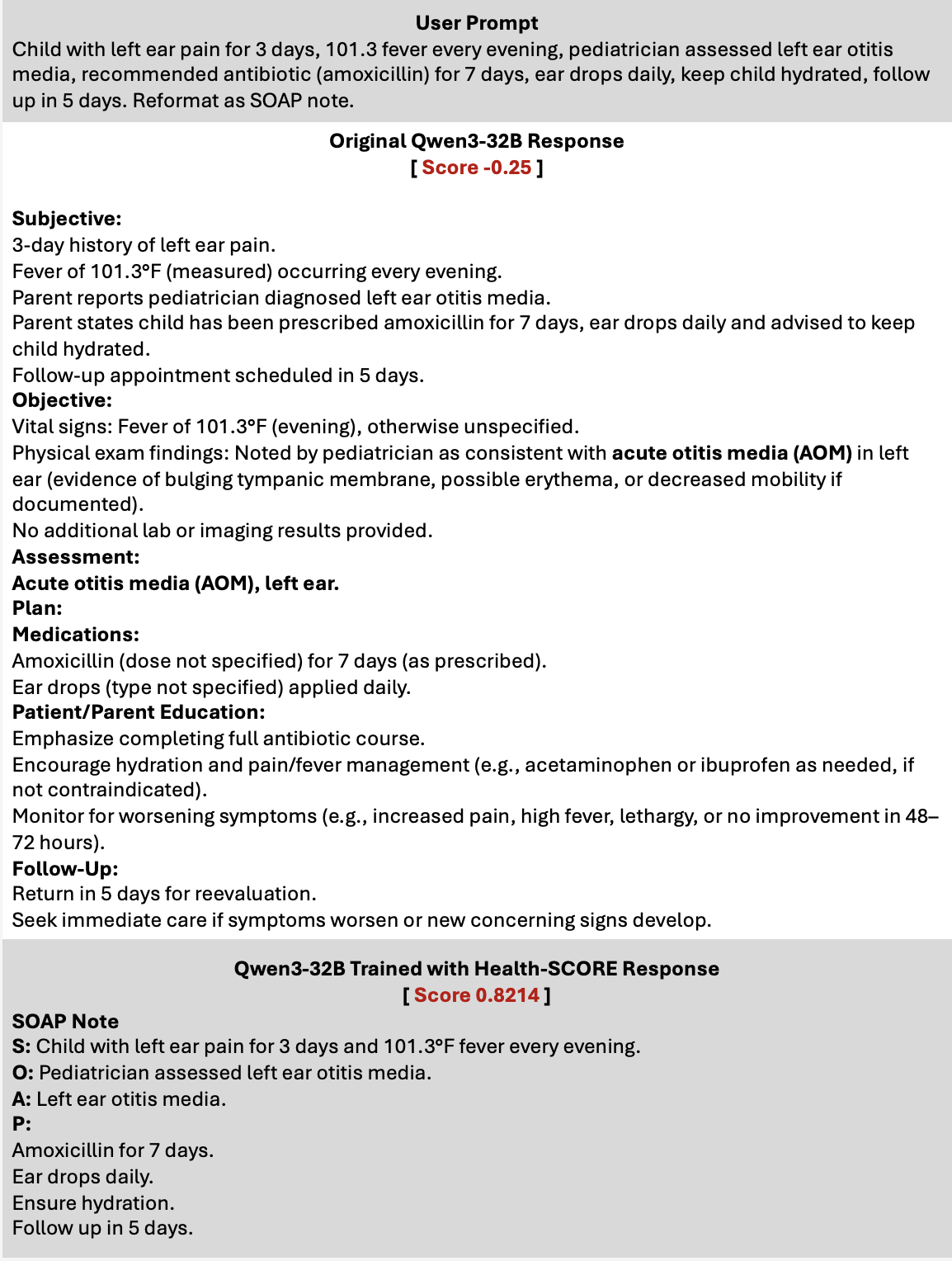}
    \caption{Qualitative example of model response.}
    \label{fig:qualitative_Example}
\end{figure*}

Figure~\ref{fig:qualitative_Example} provides a qualitative example illustrating how training with Health-SCORE leads to an improved model response. The Health-SCORE-trained model follows the SOAP format more consistently, adheres more closely to user-provided information, and produces a more accurate and concise note. In particular, it avoids introducing unsupported physical examination findings, does not speculate about the indication for the ear drops, and refrains from adding treatments that were not provided in the original input. The Health-SCORE response improves the quality of clinical documentation by producing a more appropriately structured SOAP note while remaining grounded in the available information. The note includes only the reported facts, such as the patient's symptoms, the pediatrician's assessment, amoxicillin, ear drops, hydration, and the five-day follow-up plan. In contrast, the original model response introduces unsupported physical examination findings, including a bulging tympanic membrane, erythema, and decreased mobility.

\section{Validation Guideline} 
Two domain-expert annotators from the research lab participated in the validation process and manually reviewed a subset of 500 cluster assignments. Participation was voluntary, and the annotation workload was limited. In comparison, the HealthBench paper reports that 262 physicians contributed to the benchmark over an 11-month period. For Health-SCORE, the meta-rubric validation and annotation process was completed by two domain experts over the course of several weeks, requiring substantially less annotation effort and suggesting a more scalable validation process.

\section{Use of Generative AI}  
To ensure linguistic precision and clarity, ChatGPT 5.2 was used to improve grammar, enhance readability, and assist with generating visualization code. ChatGPT was not used to generate experimental results, perform data analysis, or introduce new technical claims. All reported numbers and statements were verified by the authors against the underlying experiments and sources.

\begin{table*}[t]
\centering
\caption{Categorization of Health-SCORE Criteria}
\label{tab:healthscore_categories}
\renewcommand{\arraystretch}{1.25}

\begin{tabularx}{\textwidth}{
    |p{0.26\textwidth}
    |p{0.13\textwidth}
    |X|
}
\hline
\textbf{Category} & \textbf{Criteria} & \textbf{Description} \\
\hline

Instruction Adherence \& Response Quality
& 1, 3, 4, 5, 6, 7, 10, 12
& Instruction following, task completeness, appropriate handling of missing information and uncertainty, readability, and conciseness. \\
\hline
Information Grounding \& Internal Consistency
& 2, 25, 29
& Avoidance of fabricated information, accurate interpretation of provided information, and identification of contradictions. \\
\hline
Patient Safety, Urgency \& Critical Conditions
& 8, 9, 13, 16, 20
& Recognition of unsafe advice, clinical urgency, life-threatening conditions, and potentially important alternative conditions. \\
\hline
Diagnosis, Differential Diagnosis \& Evidence
& 19, 23, 24
& Correct diagnosis, appropriate differential diagnoses, and adherence to current clinical guidelines and evidence. \\
\hline
Testing, Imaging \& Clinical Assessment
& 18, 28
& Appropriate selection of laboratory tests, imaging, examinations, and patient-specific risk assessment. \\
\hline
Treatment \& Medication Management
& 14, 15, 17, 21, 22
& Medication safety, contraindications, antibiotic use, medication risks, combination therapy, and dose or treatment adjustment. \\
\hline
Follow-up, Referral \& Preventive Care
& 11, 26, 27
& Healthcare-provider consultation, follow-up and monitoring, preventive care, and screening recommendations. \\

\hline
\end{tabularx}
\end{table*}

\begin{figure*}[p]
    \centering
    \includegraphics[width=0.95\textwidth,height=\textheight,keepaspectratio]{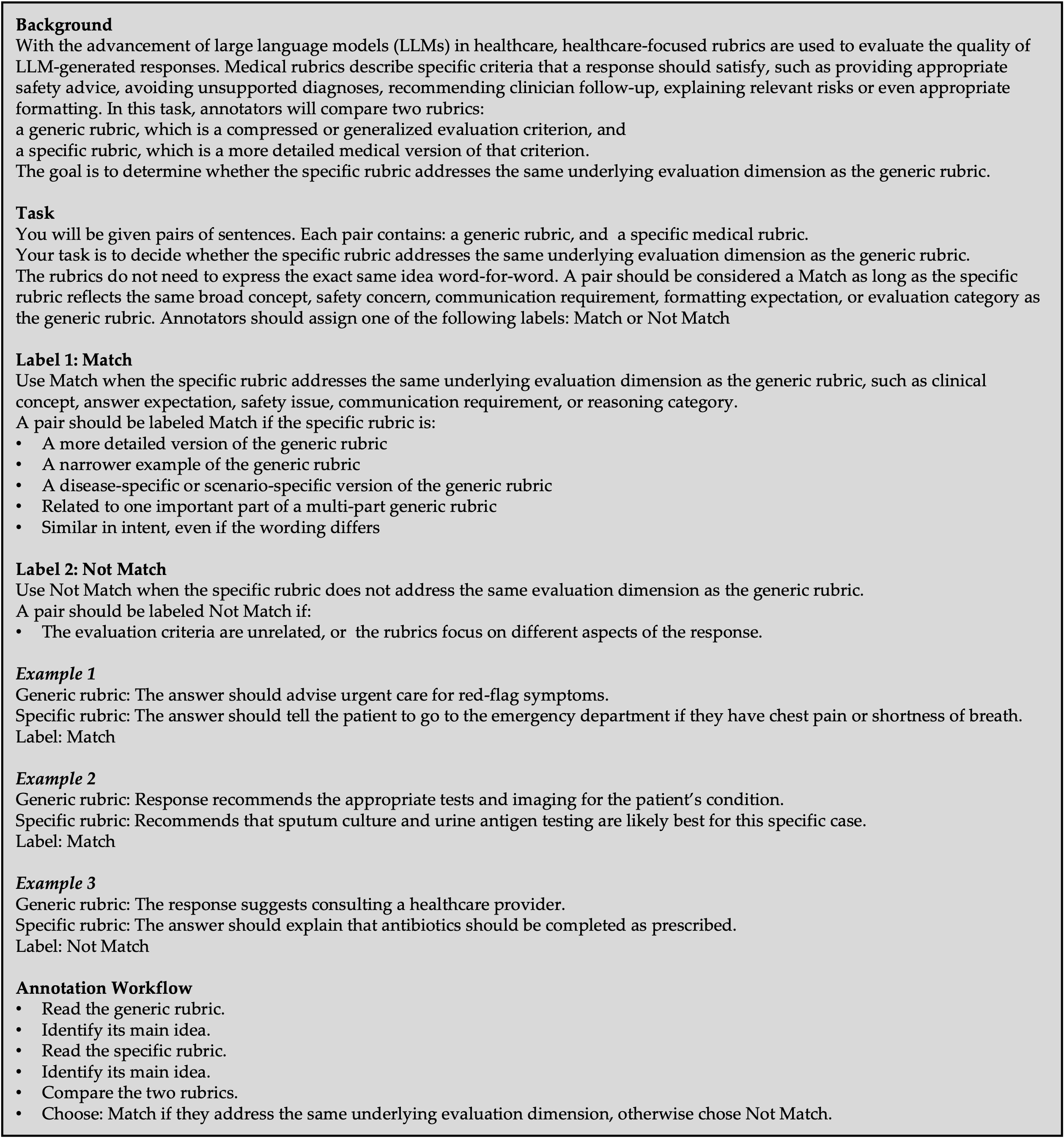}
    \caption{Validation Guideline.}
    \label{fig:annot_guideline}
\end{figure*}

\end{document}